\documentclass[final,authoryear,5p,times,twocolumn]{elsarticle}
\usepackage{amsmath}
\usepackage{amssymb} 

\usepackage{tikz}
\usetikzlibrary{arrows,matrix,positioning,patterns}
\usetikzlibrary{calc}
\usepackage{pgfplots} 
\usetikzlibrary{external}
\usetikzlibrary{angles}
\usepackage{hyperref}
\usetikzlibrary{plotmarks}

\newdefinition{rmk}{Remark}
\newproof{pf}{Proof}
\newproof{pot}{Proof of Theorem \ref{thm2}}

\usepackage[ruled, vlined, linesnumbered]{algorithm2e}

\usepackage{amsthm}

\newtheorem{problem}{Problem}

\usepackage{amsmath}
\usepackage{amssymb} 

\usepackage{eurosym}
\usepackage{colortbl}

\usetikzlibrary{arrows.meta,calc,decorations.markings,math,arrows.meta}

\usepackage[labelformat=simple]{subcaption}
\DeclareCaptionLabelSeparator{periodspace}{.\quad}
\usepackage{color,framed}
\usepackage[utf8]{inputenc} 
\usepackage{tabularx}

\usepackage{rotating}
\usepackage{graphicx}
\usepackage{lscape}
\allowdisplaybreaks 
\usepackage{longtable}

\usepackage[utf8]{inputenc}
\usepackage{pgfplots}
\usepgfplotslibrary{groupplots} 
\pgfplotsset{compat=1.3}

\makeatletter
\def\ps@pprintTitle{%
 \let\@oddhead\@empty
 \let\@evenhead\@empty
 \def\@oddfoot{}%
 \let\@evenfoot\@oddfoot}
\makeatother

\begin{document}

\begin{frontmatter}
%
%
\title{Predictive Spray Switching for an Efficient Path Planning Pattern for Area Coverage}
\author{Mogens Plessen\corref{cor1}}
\cortext[cor1]{MP is with Findklein GmbH, Switzerland, \texttt{mgplessen@gmail.com}}

%
%
%
%

\begin{abstract}
This paper presents within an arable farming context a predictive logic for the on- and off-switching of a set of nozzles. The predictive logic is tailored to a specific path planning pattern. The nozzles are assumed to be attached to a boom aligned along a working width and carried by a machinery with the purpose of applying spray along the working width. The machinery is assumed to be traveling along the specific path planning pattern. Concatenation of multiple of those path patterns and corresponding concatenation of proposed switching logics enables nominal lossless spray application for area coverage tasks. Proposed predictive switching logic is compared to the common and state-of-the-art reactive switching logic for Boustrophedon-based path planning for area coverage. The trade-off between reduction in pathlength and increase in the number of required on- and off-switchings for proposed method is discussed.
\end{abstract}
\begin{keyword}
Agriculture; Path Planning; Spraying; Predictive Switching; Overlap Avoidance.
\end{keyword}
\end{frontmatter}


\section{Introduction\label{sec_intro}}

\begin{table}
\centering
\begin{tabular}{|ll|}
\hline
\multicolumn{2}{|c|}{MAIN NOMENCLATURE}\\[2pt]
$H$ & Mainfield lane length, (m).\\
$N$ & Number of mainfield lanes, (-).\\
$R$ & Turning radius of machinery, (m).\\
$W$ & Machinery working width, (m).\\
$L_\text{path}$ & Pathlength for machinery, (m).\\
$N_\text{ON}$ & Number of nozzle switching-on states, (-).\\[2pt]
\hline
\end{tabular}
\end{table}

%
\begin{figure}
\captionsetup[subfigure]{labelformat=empty}
\centering
  \includegraphics[width=.9999\linewidth]{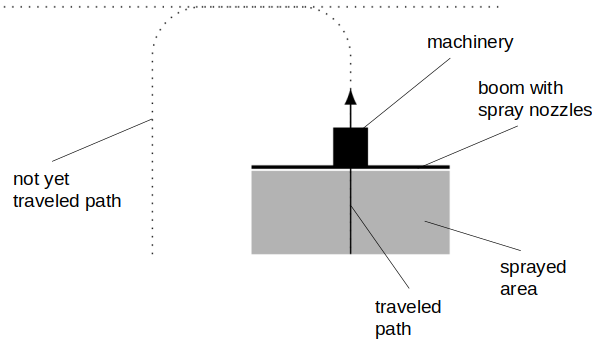}
\caption{Visualisation of two building blocks for area coverage: (i) path planning and (ii) a switching logic for the control of nozzles attached along a boom and carried by a machinery for spray application along a working width.}
\label{fig_2buildingblocks}
\end{figure}

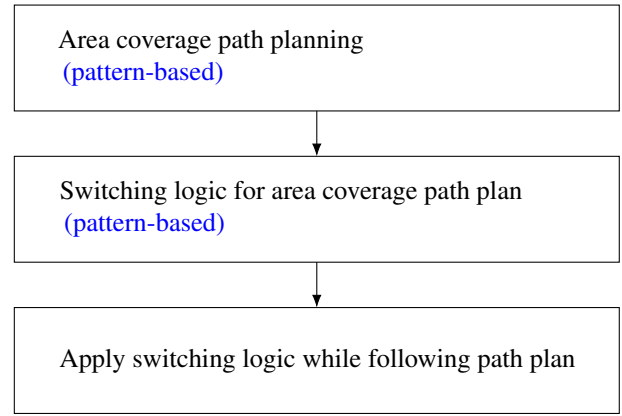
\begin{figure}
\vspace{0.3cm}
\centering%
\begin{tikzpicture}
\draw[fill=white] (0,4) rectangle (8,5.4);
\node (c) at (2.6,4.7) [scale=1,color=black,align=left] {
Area coverage path planning\\
 \hspace{0.06cm}{\color{blue}(pattern-based)}
};
\draw[black,-{Latex[scale=1.0]}] (4, 4) -- (4, 3.4);
\draw[fill=white] (0,2) rectangle (8,3.4);
\node (c) at (3.66,2.7) [scale=1,color=black,align=left] {
Switching logic for area coverage path plan\\
 \hspace{0.06cm}{\color{blue}(pattern-based)}
};
%
\draw[black,-{Latex[scale=1.0]}] (4, 2) -- (4, 1.4);
\draw[fill=white] (0,0) rectangle (8,1.4);
\node (c) at (4.01,0.7) [scale=1,color=black,align=left] {Apply switching logic while following path plan};
\end{tikzpicture}
\caption{High-level algorithm structure as a block diagram.}
\label{fig_blockdiag}
\end{figure}

Within an agricultural open-space arable farming context focusing on cereal crop cultivation of grains like wheat, rapeseed, barley and the like, area coverage applications can in general be manifold. They can include spraying, mowing, fertilizing, seeding, harvesting and so forth. 

In this paper, area coverage only relates to \emph{spraying} applications. This includes (i) spraying of herbicides, pesticides and the like for plant protection, but can alternatively (ii) also  refer to spraying of fertilizing means, or in general (iii) to applications where input means are sprayed onto a work area through one nozzle or a set of nozzles. 

Thus, methods presented in this paper do not relate to mowing, seeding and harvesting applications, or in general to applications where a machinery does not apply spray to a work area but instead operates in direct physical contact with a work area.

For this setup of area coverage planning for spraying applications there are two fundamental building blocks: (i) path planning and (ii) a switching logic for control of the nozzles, whereby those two steps follow after each other in sequence. First, a path is planned before a switching logic is applied on top, see Fig. \ref{fig_2buildingblocks} and \ref{fig_blockdiag}.

The simplest possible switching logic is to switch on at the start of the path plan and to switch off after completion of the path plan. However, within an agricultural area coverage context this may be inefficient since it typically generates spray overlap where some areas would be sprayed multiple times. Therefore, more efficient switching logics and efficient path plans are desired.

A literature review is provided for the two building blocks, path planning for area coverage and switching logics for the control of nozzles for spray application.

First, for area coverage Boustrophedon paths, whose name is derived from Ancient Greek for "like the ox turns", are mentioned. They describe a meandering or 'zig-zag' path pattern. Its usage is widespread, e.g., from search and scan applications in \cite{sequeira1979structured} to early robotic applications in \cite{tao1989coordination}. It is by far the predominant path pattern employed in practice throughout agricultural applications. The reason is that it is convenient to use by driving with agricultural machinery alternatingly one lane after another to achieve area coverage.

\begin{figure}
\centering
\begin{subfigure}[t]{.9999\linewidth}
  \centering
  \includegraphics[width=.9\linewidth]{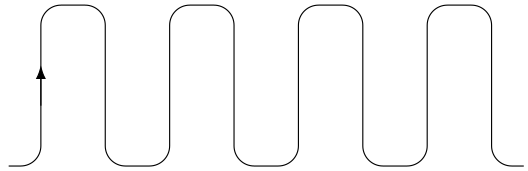}\\[-1pt]
\caption{First path pattern: Boustrophedon.\\[10pt]}
  \label{fig_prob1}
\end{subfigure}
\begin{subfigure}[t]{.9999\linewidth}
  \centering
  \includegraphics[width=.9\linewidth]{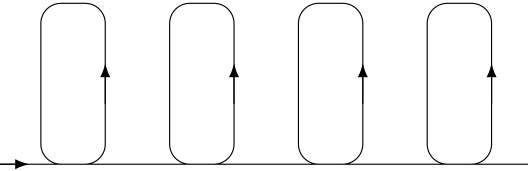}\\[-1pt]
\caption{Second path pattern: Alternative.}
  \label{fig_prob2}
\end{subfigure}
\caption{Comparison of two path patterns.}
\label{fig_2pathpatterns}
\end{figure}
%

Nevertheless, it is important to point out that this pattern is not pathlength optimal in combination with a headland path that is typical for area coverage in arable farming applications \cite{plessen2018partial}. 

The topic of optimising path planning for area coverage has been addressed by a large number of different techniques, typically tailored to the different structures of work areas. See \cite{galceran2013survey} for a 2013 survey, and \cite{tan2021comprehensive} for a more recent 2021 survey.    

\begin{figure}
\centering
\begin{subfigure}[t]{.9999\linewidth}
  \centering
  \includegraphics[width=.9\linewidth]{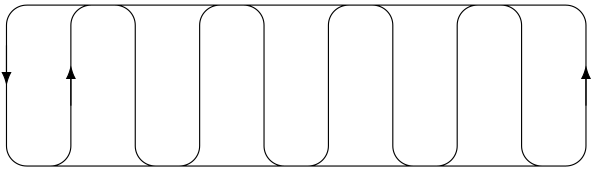}\\[-1pt]
\caption{Boustrophedon path pattern with headland path.\\[10pt]}
  \label{fig_prob1}
\end{subfigure}
\begin{subfigure}[t]{.9999\linewidth}
  \centering
  \includegraphics[width=.9\linewidth]{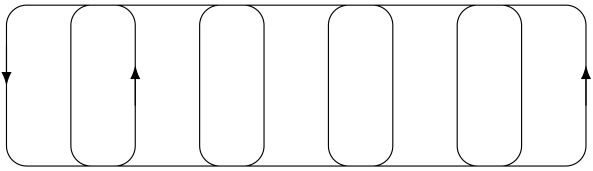}\\[-1pt]
\caption{Alternative path pattern with headland path.}
  \label{fig_prob2}
\end{subfigure}
\caption{Comparison of two path patterns with a headland path.}
\label{fig_2pathpatternswheadl}
\end{figure}

Within the arable farming context the inclusion of a headland path is a characteristic constraint that has to be taken into account for area coverage path planning (see e.g. \cite{he2023dynamic, utamima2022agricultural, pour2023complete, hoffmann2023optimal, jayalakshmi2025comprehensive, asiminari2024integrated, hoffmann2024optimal}). For illustration, compare Fig. \ref{fig_2pathpatterns} and Fig. \ref{fig_2pathpatternswheadl} for area coverage paths with and without headland paths, respectively. Headland paths are required for full area coverage when operating nonholonomic vehicles such as tractors with a limited turning radius within the work area. 

In \cite{plessen2019optimal} optimal in-field routing was discussed for  arbitrary non-convex fields and multiple obstacle areas. In general, in such a setup the optimal solution can result in a route that would be unintuitive to drive in contrast to a Boustrophedon path. Nevertheless, it was found that there exists a specific path planning \emph{pattern} that often can form part of an optimal solution. This path pattern was further analyzed in \cite{plessen2018partial}. This pattern shall also represent the pattern based on which a predictive spray switching logic is presented in this paper.

The second building block is discussed. Spraying, which is typically applied to the work area via a set of nozzles aligned along a boom for broadcast spraying\footnote{The term \emph{broadcast spraying} implies that spray is applied over the entire width of the boom. For broadcast spraying nozzles are spaced along the boom such that individual nozzle sprays overlap such that uniform spray coverage along the entire boom width can be achieved.} \cite{portman1979calibrating, smith2000broadcast, vijayakumar2023smart}, is a highly dynamic process and affected by a plethora of parameters. These include nozzle type, spray fan angle, spray pressure, boom height, nozzle spray overlap, nozzle spacing, nozzle clogging, machinery traveling speed, cross wind for spray drift, terrain undulations and more \cite{holterman1997modelling, hassen2013effect, mangus2017analyzing, burgers2021comparison, al2023development, wang2023evaluation, wang2023evaluationof,  li2023development, saleem2023variable}. For unmanned aerial spraying \cite{carreno2022numerical}, in contrast to traditional spraying with tractors carrying or trailing spraying machinery and sprayers operating close to the ground, dynamic effects are further amplified.

Aforementioned effects are listed to emphasize the high real-world complexity of the spraying process. For the remainder of this paper, however, (i) \emph{nominal} instant switching and (ii) absence of any of above mentioned spray dynamics and spray transients are assumed. This is justified for two reasons: First, spray effects hold simultaneously for \emph{both} the Boustrophedon-based path pattern and the suggested alternative path pattern. Second, for a deterministic planning problem this paper presents a novel \emph{deterministic} solution technique, i.e., a predictive deterministic switching logic that exploits the structure of a specific path planning pattern. For this presentation spray transients are omitted in the following since these do not alter the general switching logic. Therefore, in the following the two objective evaluation metrics (i) pathlength and (ii) the number of on/off-switches are used to compare the proposed predictive method to an alternative reactive Boustrophedon-based and state-of-the-art baseline under nominal conditions. Analytical formulas are derived that underline the potential of the proposed method for pathlength savings that scale linearly with the number of mainfield lanes and the working width.

The research gap and motivation for this paper is discussed. There is a research gap in linking spray switching logics to path planning patterns that are different from the Boustrophedon pattern. Moreover and in particular, individual nozzle control is typically applied \emph{reactively} varying laterally along the boom withtin the framework of variable rate automatic section (boom section or nozzle) control \cite{luck2010potential, sharda2011boom}. The reactive aspect implies that nozzles are switched on during path traversal when an area has not yet been sprayed and switched off (i) when an area is traversed a second or more times such that no overlap occurs with already sprayed areas and (ii) when an area is traversed that is not meant to be sprayed according to a map \cite{luck2010potential}. 

In contrast, to the best of the author's knowledge this is the first paper that proposes (i) a \emph{predictive} switching logic that (ii) exploits the structure of an efficient path planning pattern for area coverage that is different from the Boustrophedon pattern.

%

The remaining paper is organised as follows: problem formulation, problem solution, numerical results and the conclusion are described in Sections \ref{sec_probformulation}-\ref{sec_conclusion}.

\section{Problem Formulation\label{sec_probformulation}}

The problem addressed in this paper is as follows:

\begin{problem}\label{problem1}
Given a convexly-shaped two-dimensional work area, determine a path plan for area coverage based on the concatenation of a recurring path pattern and determine a corresponding logic for the on- and off-switching of a set of nozzles attached to a boom aligned along the working width. The boom with nozzles shall be carried by a machinery or trailed by a machinery as an implement with the purpose of applying spray along the working width while the machinery is traveling along the path plan, subject to the constraint that the path plan for area coverage shall include a headland path.
\end{problem}

Three comments are made. First, the constraint of inclusion of a headland path is a typical setup for outdoors agricultural applications. It is warranted (i) for nonholonomic vehicles such as tractors that typically operate in agricultural fields with a limited turning radius, and (ii) in order to ensure field contours shall not be exceeded or violated while still enabling full area coverage. Note that the benefits of inclusion of a headland path can also apply to drone applications \cite{plessen2025path}. Second, the assumption of a convexly-shaped work area enables that Problem \ref{problem1} can be fully solved by the concatenation of a recurring path pattern. Third, the treatment of non-convexly shaped work areas and work areas that further include obstacle areas such as tree islands, ponds, power pole masts and so forth is more complex and not subject of this short paper, where the focus is on presentation of a switching logic for a specific alternative path planning pattern. However, short comments and an outlook for the non-convex setup are provided at the end of Section \ref{sec_solnMethod}.

\section{Problem Solution\label{sec_solnMethod}}

This section discusses two solution approaches for Problem \ref{problem1}. First, the state-of-the-art and widespread method applied in agricultural practice that is based on Boustrophedon path planning and a corresponding reactive switching logic is discussed. Second, an alternative proposition that is based on an alternative path pattern and a predictive switching logic is presented. The high-level approach is illustrated in Fig. \ref{fig_blockdiag}.

\subsection{Boustrophedon-based Reactive Switching Method\label{subsec_Boustro}}

\begin{figure}
\centering
\begin{subfigure}[t]{.9999\linewidth}
  \centering
  \includegraphics[width=.94\linewidth]{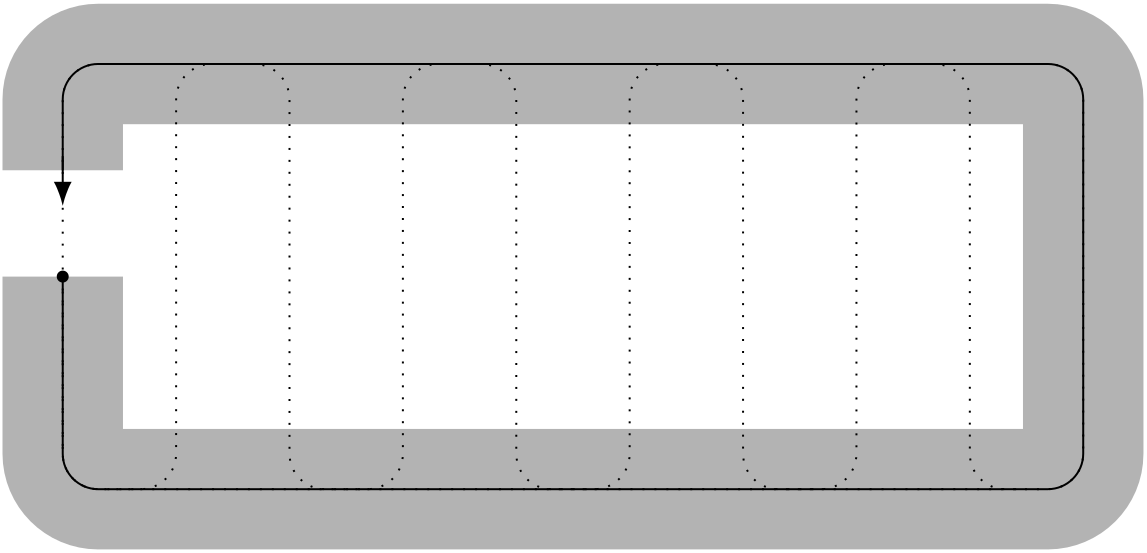}\\[-1pt]
\caption{Initial full headland path coverage.\\[10pt]}
  \label{fig_Boustro_step1}
\end{subfigure}
\begin{subfigure}[t]{.9999\linewidth}
  \centering
  \includegraphics[width=.94\linewidth]{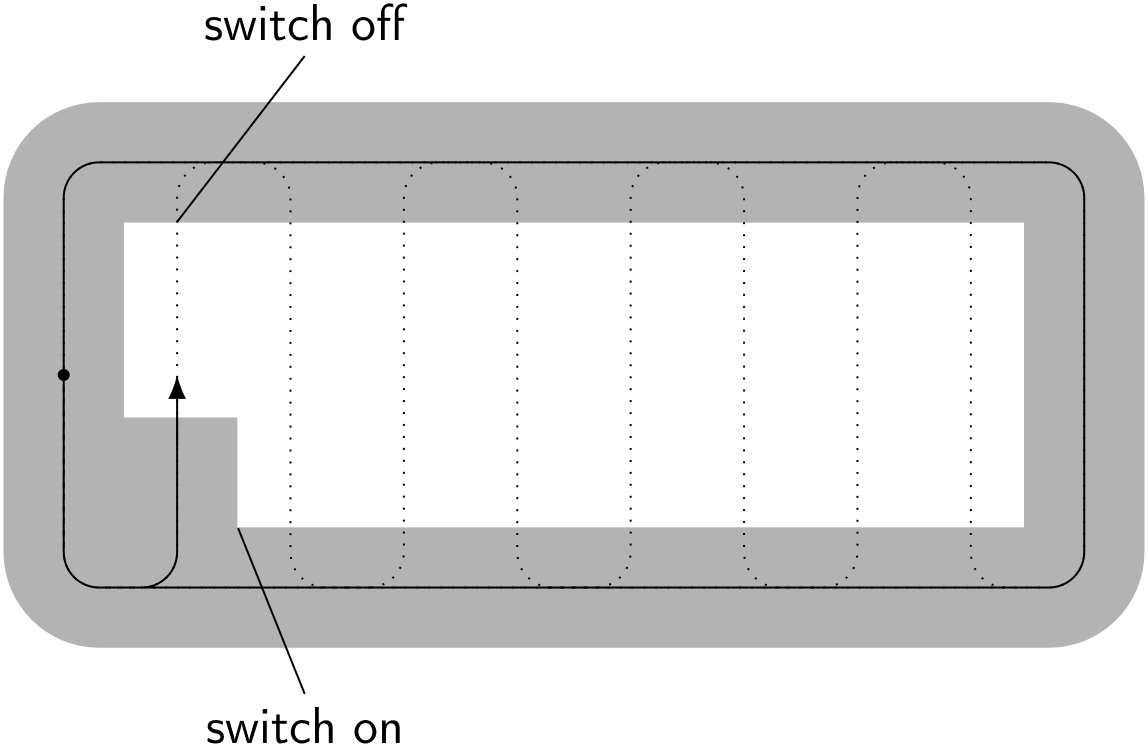}\\[1pt]
\caption{Reactive switching along mainfield lanes.\\[10pt]}
  \label{fig_Boustro_step2}
\end{subfigure}
\begin{subfigure}[t]{.9999\linewidth}
  \centering
  \includegraphics[width=.94\linewidth]{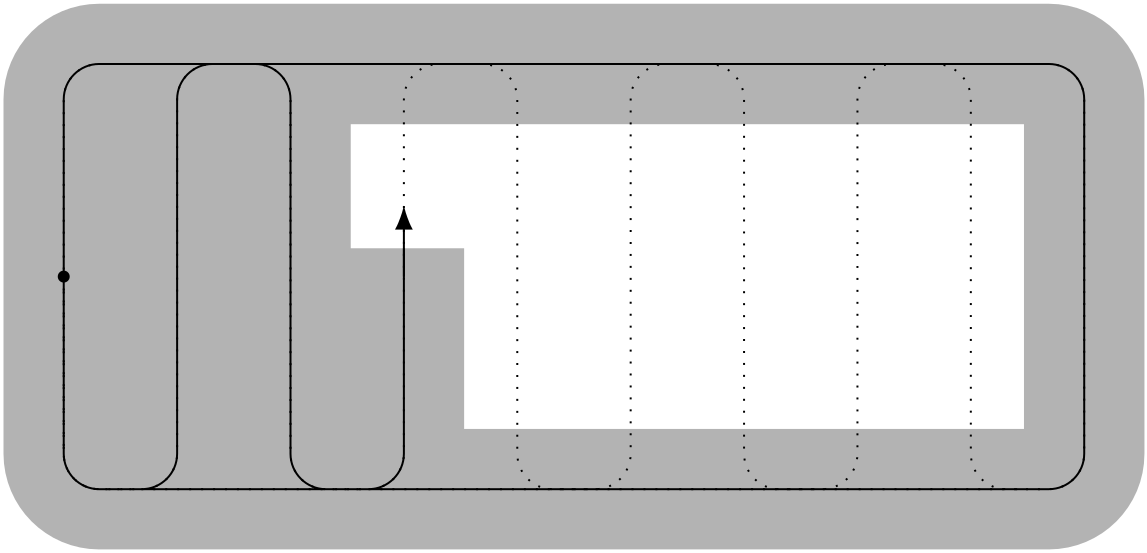}\\[-1pt]
\caption{Reactive switching along mainfield lanes.\\[10pt]}
  \label{fig_Boustro_step3}
\end{subfigure}
\begin{subfigure}[t]{.9999\linewidth}
  \centering
  \includegraphics[width=.94\linewidth]{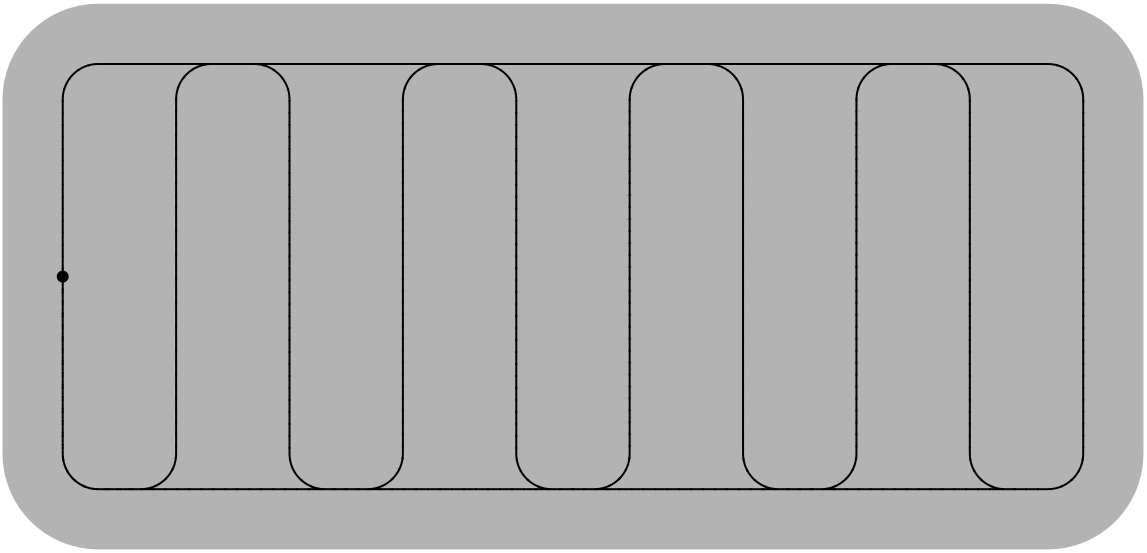}\\[-1pt]
\caption{Result after full traversal of the path plan.}
  \label{fig_Boustro_step4}
\end{subfigure}
\caption{Method 1: State-of-the-art reactive switching logic for area coverage based on the Boustrophedon-path pattern in combination with an initial full headland path traversal. The field entrance and starting point of the path is indicated by the black dot.}
\label{fig_Boustro_4steps}
\end{figure}

The switching logic for Boustrophedon-based path planning for area coverage is visualised in Fig. \ref{fig_Boustro_4steps}. It is switched on along the initial full headland path traversal beginning at the field entrance, see Fig. \ref{fig_Boustro_step1}. Then, it is \emph{reactively} switched on along traversals of mainfield lanes as soon as a work area has not yet been sprayed. This occurs along the mainfield lanes at closest projection distances of half the operating width, $W/2$, away from the headland path. Likewise, it is reactively switched off towards the end of traversal of a mainfield lane at closest projection distances of half the operating width, $W/2$, away from the headland path such that no overlap occurs with already sprayed area. See Fig. \ref{fig_Boustro_step2} for illustration. This procedure is repeated while the machinery is traversing the mainfield lanes in the typical Boustrophedon pattern, see Fig. \ref{fig_Boustro_step3}. After traversal of the last mainfield lane the machinery travels along the shortest path along the headland path to the field exit with spray nozzles switched off such as to avoid spray overlap. The final result of this reactive spraying method implies full area coverage, see Fig. \ref{fig_Boustro_step4}.

%

\subsection{Alternative Predictive Switching Method\label{subsec_alternative}}

%
\begin{figure}
\captionsetup[subfigure]{labelformat=empty}
\centering
  \includegraphics[width=.8\linewidth]{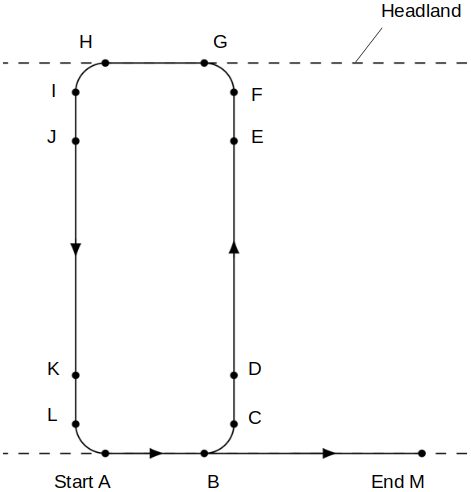}
\caption{Method 2: An instance of the path planning pattern of interest and relevant waypoints along its path.}
\label{fig_1pattern}
\end{figure}

\begin{figure}
\centering
\begin{subfigure}[t]{.49\linewidth}
  \centering
  \includegraphics[width=.96\linewidth]{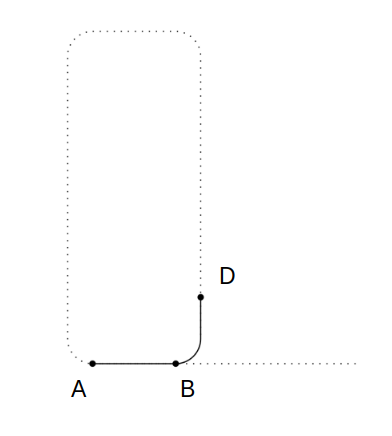}\\[-1pt]
\caption{\\[10pt]}
  \label{fig_step1_switchinglogic}
\end{subfigure}
~\begin{subfigure}[t]{.49\linewidth}
  \centering
  \includegraphics[width=.96\linewidth]{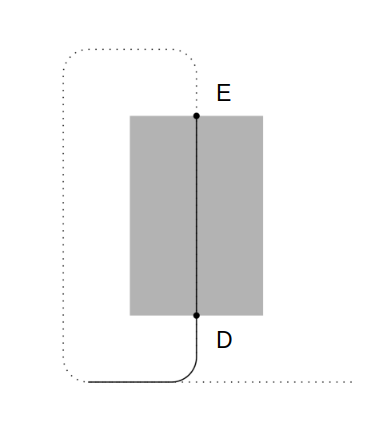}\\[-1pt]
\caption{\\[10pt]}
  \label{fig_step2_switchinglogic}
\end{subfigure}
\begin{subfigure}[t]{.49\linewidth}
  \centering
  \includegraphics[width=.96\linewidth]{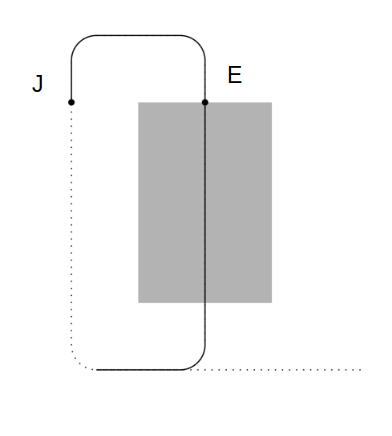}\\[-1pt]
\caption{\\[10pt]}
  \label{fig_step3_switchinglogic}
\end{subfigure}
\begin{subfigure}[t]{.49\linewidth}
  \centering
  \includegraphics[width=.96\linewidth]{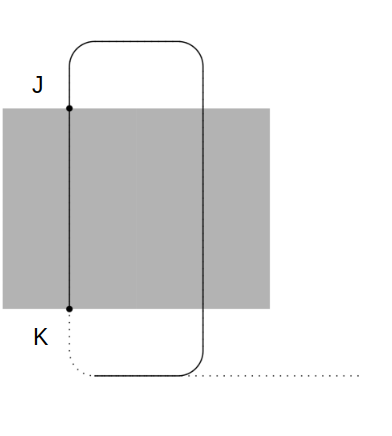}\\[-1pt]
\caption{\\[10pt]}
  \label{fig_step4_switchinglogic}
\end{subfigure}
\begin{subfigure}[t]{.49\linewidth}
  \centering
  \includegraphics[width=.96\linewidth]{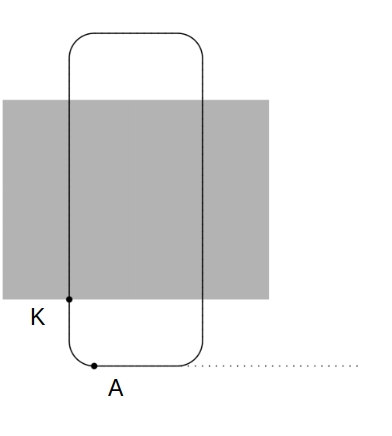}\\[-1pt]
\caption{\\[10pt]}
  \label{fig_step5_switchinglogic}
\end{subfigure}
\begin{subfigure}[t]{.49\linewidth}
  \centering
  \includegraphics[width=.96\linewidth]{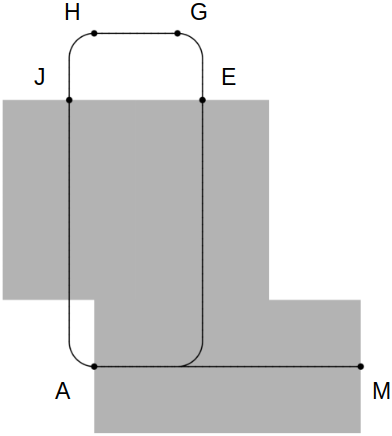}\\[-1pt]
\caption{\\[10pt]}
  \label{fig_step6_switchinglogic}
\end{subfigure}
\caption{Method 2: Illustration of proposed predictive switching logic (a)-(f) for the path pattern in Fig. \ref{fig_1pattern}. Gray areas indicate sprayed area. See Section \ref{subsec_alternative} for a description.}
\label{fig_6steps_switchinglogic}
\end{figure}
%

The second method to address Problem \ref{problem1} differs from the previous Boustrophedon-based reactive method in two ways: (i) a different path pattern is employed, and (ii) the switching method includes a predictive logic. 

The path pattern and its waypoints of interest are highlighted in Fig. \ref{fig_1pattern}. Several comments are made. First, the path traversal begins a start point A and follows the letters in order A, B, C, \dots ~until end point M. Second, path segments B-C, F-G, H-I and L-A indicate turn maneuvers for transitions between headland path and mainfield lanes. A turning radius $R>0$ for nonholonomic vehicle dynamics is assumed. Third, waypoints D, E, J and K indicate locations along mainfield lanes that have a closest projection distance of half the working width, $W/2$, away from the headland path. Fourth, headland path segments part of the path pattern are segments A-B, G-H and A-M. Note that the initial segment A-B is also a sub-segment of the segment A-M.

\newpage

For the path pattern in Fig. \ref{fig_1pattern} the proposed switching logic is visualised in Fig. \ref{fig_6steps_switchinglogic} and is as follows: 
\begin{itemize}
\item[(a)] Along transition A-D it is switched off.
\item[(b)] Along transition D-E it is switched on.
\item[(c)] Along transition E-J it is switched off.
\item[(d)] Along transition J-K it is switched on.
\item[(e)] Along transition K-A it is switched off.
\item[(f)] Along transition A-M it is switched on.
\end{itemize}
Multiple comments are made. First, according to the logic it is never switched on along turn maneuvers for the transition between headland path and mainfield lanes, see Steps (a), (c) and (e). This is beneficial in that during such turn maneuvers nozzles located at different locations along the boom are exhibiting different traveling velocities. In order to maintain uniform spray application along the entire working width in such scenarios individual nozzle control \cite{luck2010potential} would be required. This is here avoided.

%
\begin{figure}
\captionsetup[subfigure]{labelformat=empty}
\centering
  \includegraphics[width=.82\linewidth]{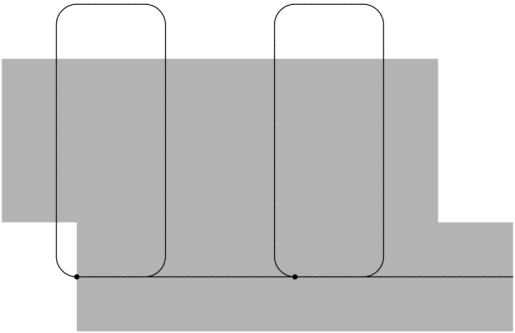}
\caption{Method 2: Concatenation of two path planning patterns.}
\label{fig_concat2patterns}
\end{figure}

Second, according to Step (b) and (d) it is switched on only along segments of mainfield lanes that are in closest projection distance at least half the operating width away from the headland path. This is because the area around segments of the mainfield lanes that are close to the headland path are more efficiently sprayed during traversal of the headland path segments. This avoids spraying during turn maneuvers for transitions between headland path and mainfield lanes.

Third, assuming nominal instant switching and spray application without a transient phase\footnote{This is a hypothetical and ideal setup used throughout this paper to present the method and to analyse its theoretical characteristics. See also the discussion in the Introduction and in Sect. \ref{subsec_noncvx}.} there is zero overlap in the total sprayed area as shown in Fig. \ref{fig_step6_switchinglogic} indicated by the gray area. This is the result of the three switching-on phases illustrated in Fig. \ref{fig_6steps_switchinglogic}.

Finally, the two predictive characteristics of the switching pattern are discussed. The first predictive aspect of the method involves the headland path segment A-B in Fig. \ref{fig_1pattern}. The first transition A-D according to Fig. \ref{fig_step1_switchinglogic} is traversed with switching-off state. This transition includes traversal of the path segment A-B along the headland path. According to above switching logic it is explicitly switched off along this transition. This is since by knowledge of the path pattern it is predictively known that the path segment A-B will be traversed a \emph{second} time as part of the sixth and final transition A-M of the path pattern. Importantly, the entire transition A-M, which includes the path segment A-B, is along the headland path. In contrast, the first ttransition A-D, which also includes the path segment A-B, is only partly along the headland path, but also partly along a turn maneuver for the transition from headland path to mainfield lane. Note that such a predictive switching logic is absent from the state-of-the-art reactive switching logic for Boustrophedon-based area coverage path planning described in Sect. \ref{subsec_Boustro}. 

The second predictive aspect of the method involves the concatenation of multiple path patterns and their switching logics. As the sprayed area in Fig. \ref{fig_step6_switchinglogic} illustrates, after the traversal of the path pattern and according to the switching logic outlined in Fig. \ref{fig_step1_switchinglogic}-\ref{fig_step6_switchinglogic} not all segments along the path pattern are sprayed during their traversal. In particular, this relates to the transition E-J in Fig. \ref{fig_step3_switchinglogic} and transition K-A in Fig. \ref{fig_step5_switchinglogic}. However, being  aware of the embedding of the path pattern within an overall area coverage path plan the coverage of these missing segments can be achieved by:
\begin{itemize}
\item[(i)] Concatenating multiple path patterns.
\item[(ii)] Introducing a switching logic for the initial transition from a field entrance to the first occurrence of the path pattern according to the overall area coverage path plan.
\item[(iii)] Introducing a switching logic for the path after completion of the last path pattern according to the overall area coverage path plan. 
\end{itemize}
These three aspects are elaborated on next.

\begin{figure}
\centering
\begin{subfigure}[t]{.9\linewidth}
  \centering
  \includegraphics[width=.7\linewidth]{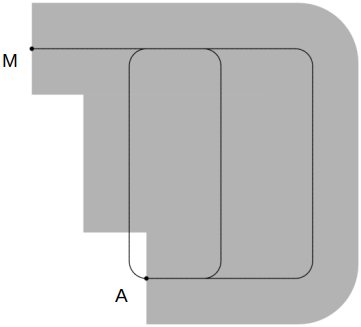}\\[-1pt]
\caption{First special case example of a prolonged headland path segment.\\[10pt]}
  \label{fig_2specialcases_1}
\end{subfigure}
\begin{subfigure}[t]{.9\linewidth}
  \centering
  \includegraphics[width=.7\linewidth]{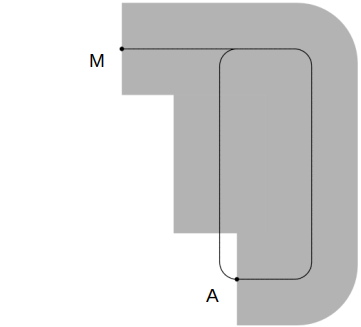}\\[-1pt]
\caption{Second special case example of a prolonged headland path segment, whereby this segment is furthermore replacing the second mainfield lane that is otherwise typical for the path pattern in Fig. \ref{fig_1pattern}.}
  \label{fig_2specialcases_2}
\end{subfigure}
\caption{Method 2: Two special cases with prolonged headland path segments.}
\label{fig_2specialcases}
\end{figure}
%

\begin{figure}
\captionsetup[subfigure]{labelformat=empty}
\includegraphics[width=5.5cm]{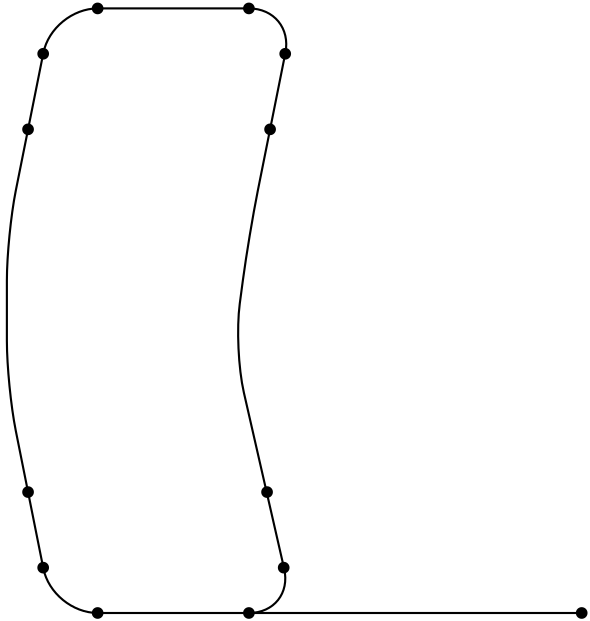}
\caption{The proposed switching logic is also valid for freeform mainfield lanes. Relevant waypoints remain conceptually as in Fig. \ref{fig_1pattern}.}
\label{fig_curve}
\end{figure}

\begin{figure}
\centering
\begin{subfigure}[t]{.9999\linewidth}
  \centering
  \includegraphics[width=.99\linewidth]{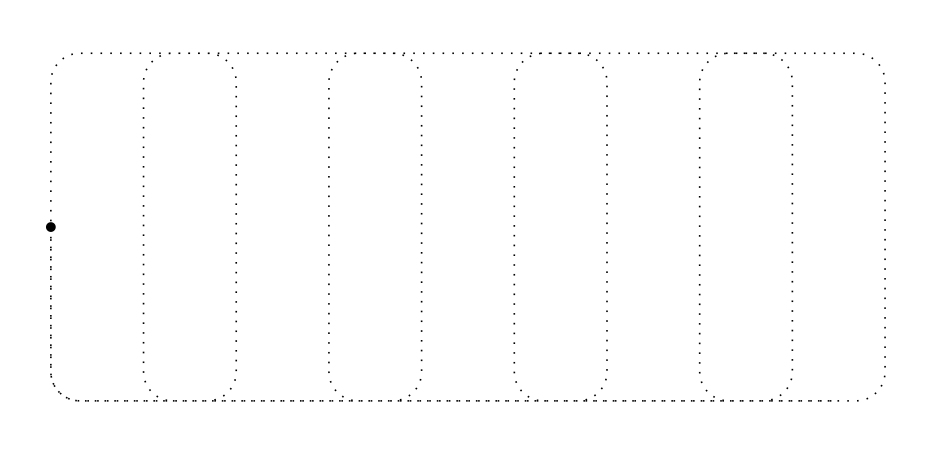}\\[-1pt]
\caption{Planned area coverage path.\\[10pt]}
  \label{fig_prob1}
\end{subfigure}
\begin{subfigure}[t]{.9999\linewidth}
  \centering
  \includegraphics[width=.99\linewidth]{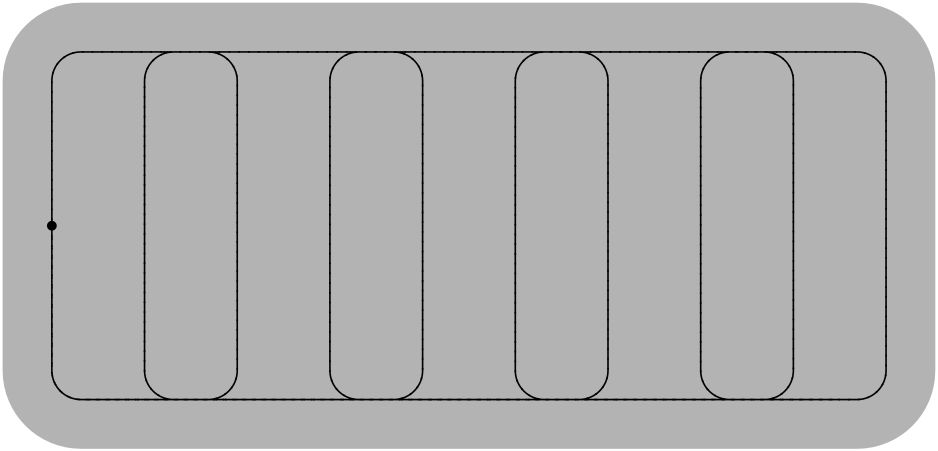}\\[-1pt]
\caption{Spraying result after traversal of the planned area coverage path and application of proposed switching logic.}
  \label{fig_prob2}
\end{subfigure}
\caption{Method 2: Visualisation of a full area coverage example resulting from the concatenation of multiple path patterns, application of proposed switching logic in Fig. \ref{fig_6steps_switchinglogic}, and special cases handling according to Fig. \ref{fig_2specialcases}.}
\label{fig_fullareacovg}
\end{figure}
%

Regarding (i), Fig. \ref{fig_concat2patterns} illustrates the concatenation of two path patterns, illustrating how a concatenation enables spraying of a larger area.

Regarding (ii), the initial switching logic for the initial transition from a field entrance to the first occurrence of the path pattern shall be \emph{reactive}. Thus, during this phase it shall be switched on during path traversal when an area has not yet been sprayed and switched off when an area is traversed a second or more times. 

Regarding (iii), the switching logic for the path after completion of the last path pattern according to overall area coverage path plan shall likewise be \emph{reactive}. Thus, it shall be switched on during path traversal when an area has not yet been sprayed and switched off when an area is traversed a second or more times.

Finally, Fig. \ref{fig_2specialcases} illustrates two special cases of the path pattern in Fig. \ref{fig_1pattern}, where a headland path segment is prolonged. Two comments are made.

First, if an area coverage path plan consists entirely of a concatenation of path patterns of Fig. \ref{fig_1pattern} then one of those special cases marks the last path pattern according to the overall area coverage path plan. Then, waypoint M in Fig. \ref{fig_2specialcases} marks the field exit point, whereby in general the final headland path segment A-M can span multiple path patterns.

Second, it is noted that Fig. \ref{fig_2specialcases_1} and Fig. \ref{fig_2specialcases_2} represent boundary cases. In general, any intermediate case is also possible. In  such scenarios, spray overlap occurs resulting from spraying along the prolonged headland path segment and one mainfield lane. However, this can be mitigated by boom section control \cite{luck2010potential,sharda2011boom} and does not affect the general switching logic. The effect of spray overlap for the area between the first or last mainfield lane and the headland path is a well-known phenomenon in agriculture, and does also occur for the Boustrophedon-based path planning and switching logic.

For a full area coverage example illustrating above rules (i)-(iii) see Fig. \ref{fig_fullareacovg}.

\subsection{Non-convexly shaped work areas\label{subsec_noncvx}}

As outlined in Problem \ref{problem1} the scope of this paper are convexly shaped work areas. The objective of this paper is presentation of an efficient predictive switching logic for a specific path planning pattern.

Not subject of this paper is the application of the method to non-convexly shaped work areas. Here, optimised area coverage path planning (e.g. \cite{plessen2019optimal}) is more complex and area coverage paths often consist of concatenations of the path pattern in Fig. \ref{fig_1pattern} but often also consists of freely optimised and less intuitive routing paths. Above switching logics, and in particular listing (i)-(iii), are still maintained. Likewise, also for the case of field contours described by polygons and freeform mainfield lanes (see e.g. \cite{plessen2021freeform} and Fig. \ref{fig_curve}) switching logics according to Sect. \ref{subsec_alternative} are still maintained.

The evaluation of the method for (i) non-convexly shaped real-world work areas, plus (ii) the inclusion of turn compensation for individual boom section control, and (iii) the effect of spray transients during switching is subject of ongoing work.

\section{Numerical Results and Discussion\label{sec_IllustrativeEx}}


%
\begin{figure}
\captionsetup[subfigure]{labelformat=empty}
\centering
  \includegraphics[width=.99\linewidth]{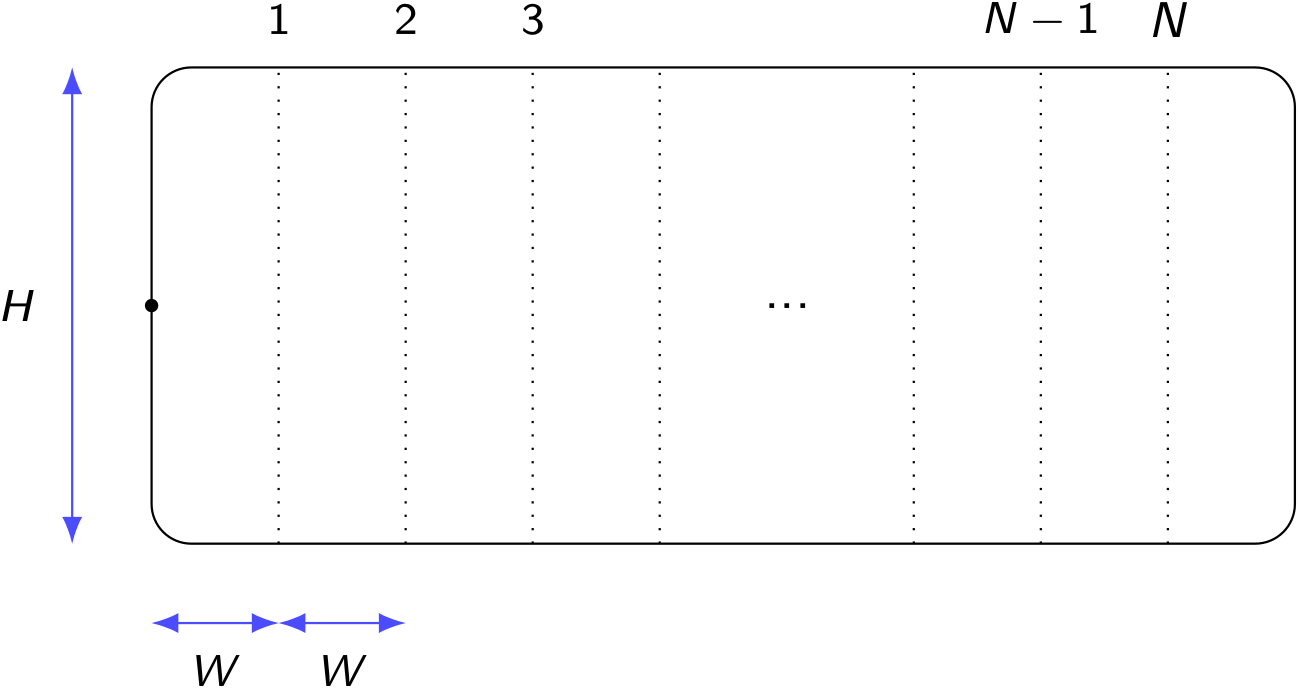}
\caption{Experiments setup: Two path planning methods and their switching logics are compared in two experimental setups, one for an even and one for an odd number of $N>0$ mainfield lanes, as a function of variable $N$. The working width is $W>0$. The uniform mainfield lane length is $H>0$. A turning radius $R>0$ is assumed. The typical curved trajectories, that would occur at the transitions between mainfield lanes and headland path for nonholonomic vehicles, are not displayed in above plot, since they vary for different path plans according to the Boustrophedon-based or Alternative method. They are, however, accounted for in the calculations in \eqref{eq_oddN} and \eqref{eq_evenN}.}
\label{fig_exptsetup}
\end{figure}

\begin{figure*}
\vspace*{-1cm}\centering
\begin{subfigure}[t]{.9999\linewidth}
  \centering
  \includegraphics[width=.37\linewidth]{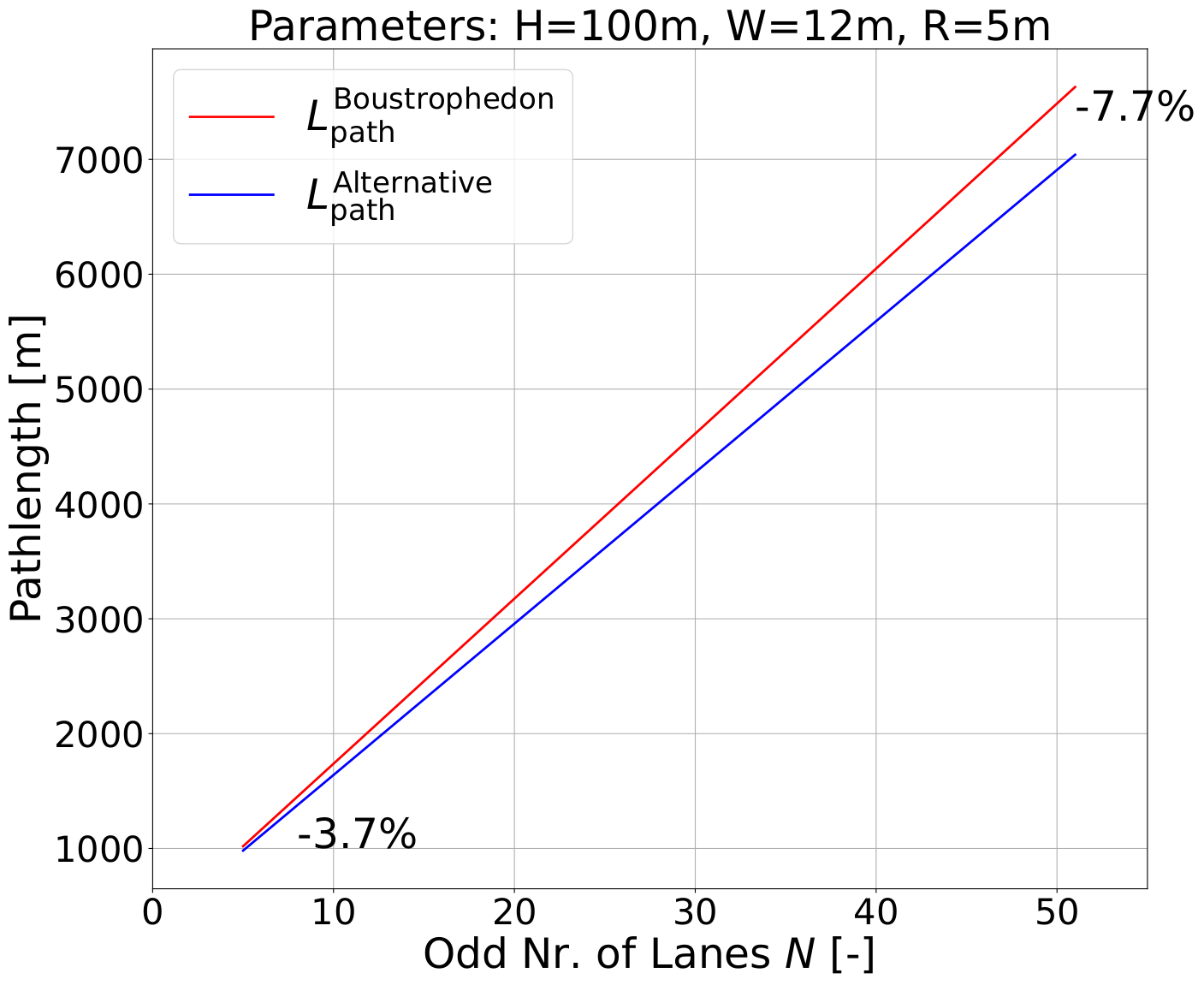}~
  \includegraphics[width=.37\linewidth]{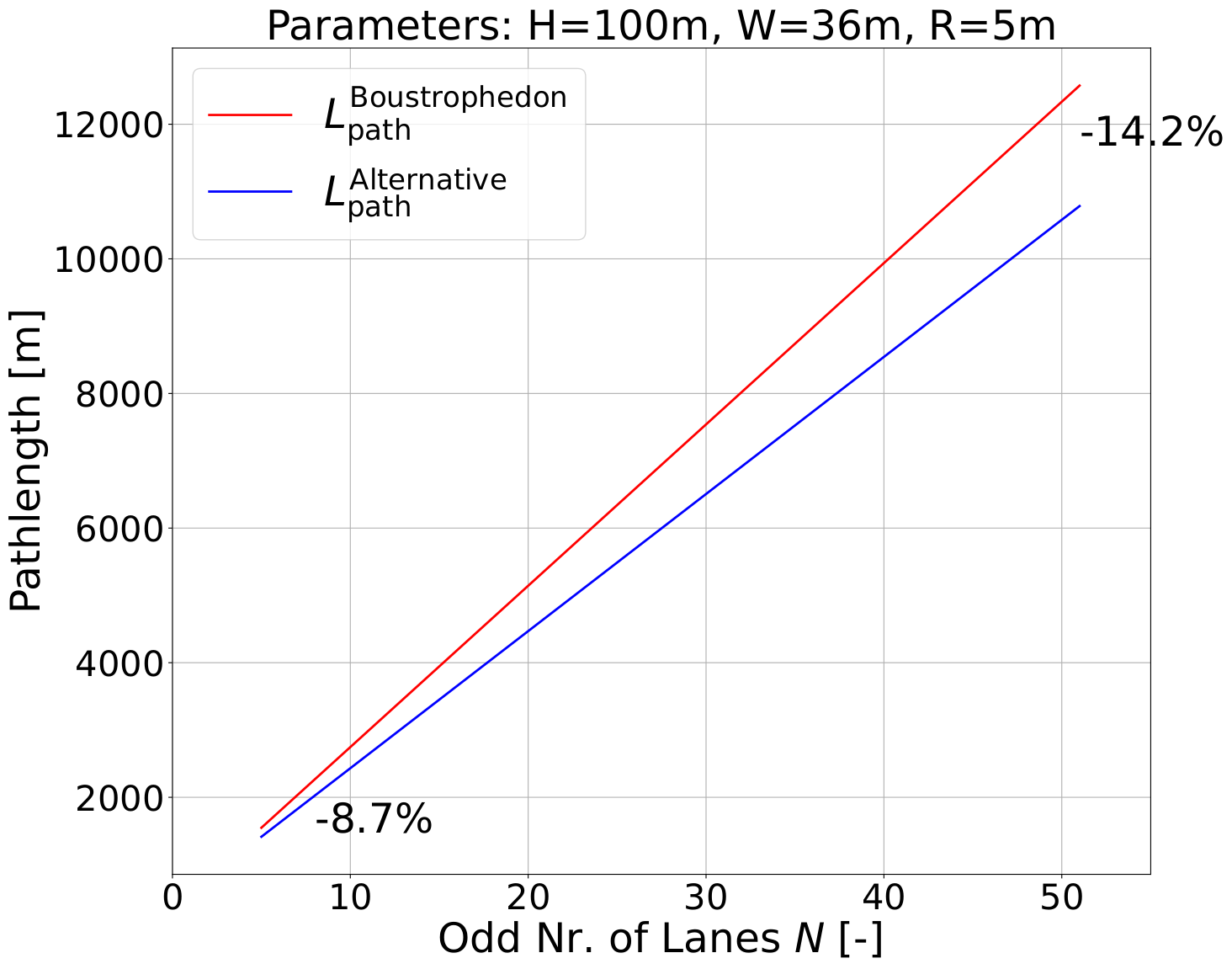}\\[-1pt]
  \includegraphics[width=.37\linewidth]{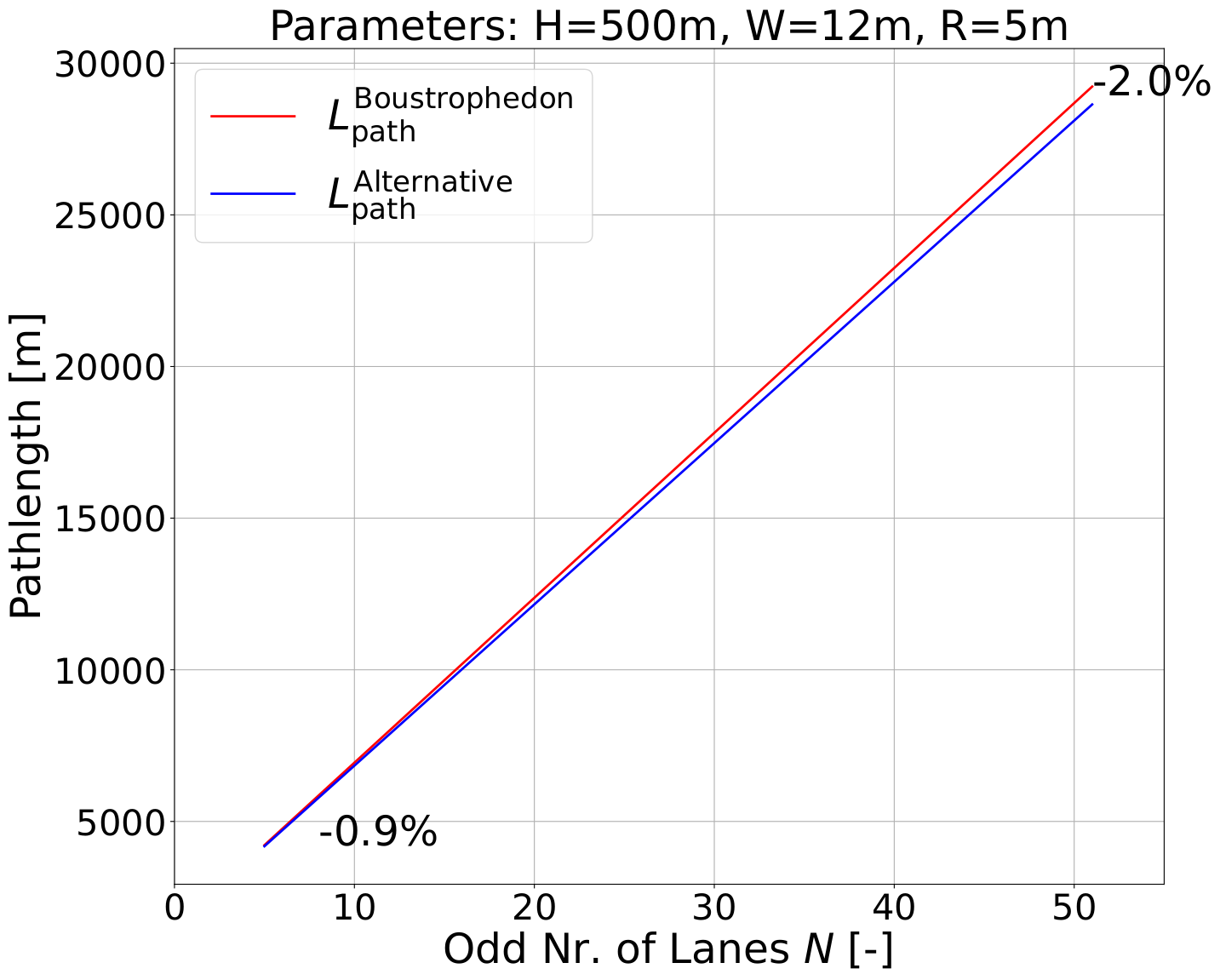}~
  \includegraphics[width=.37\linewidth]{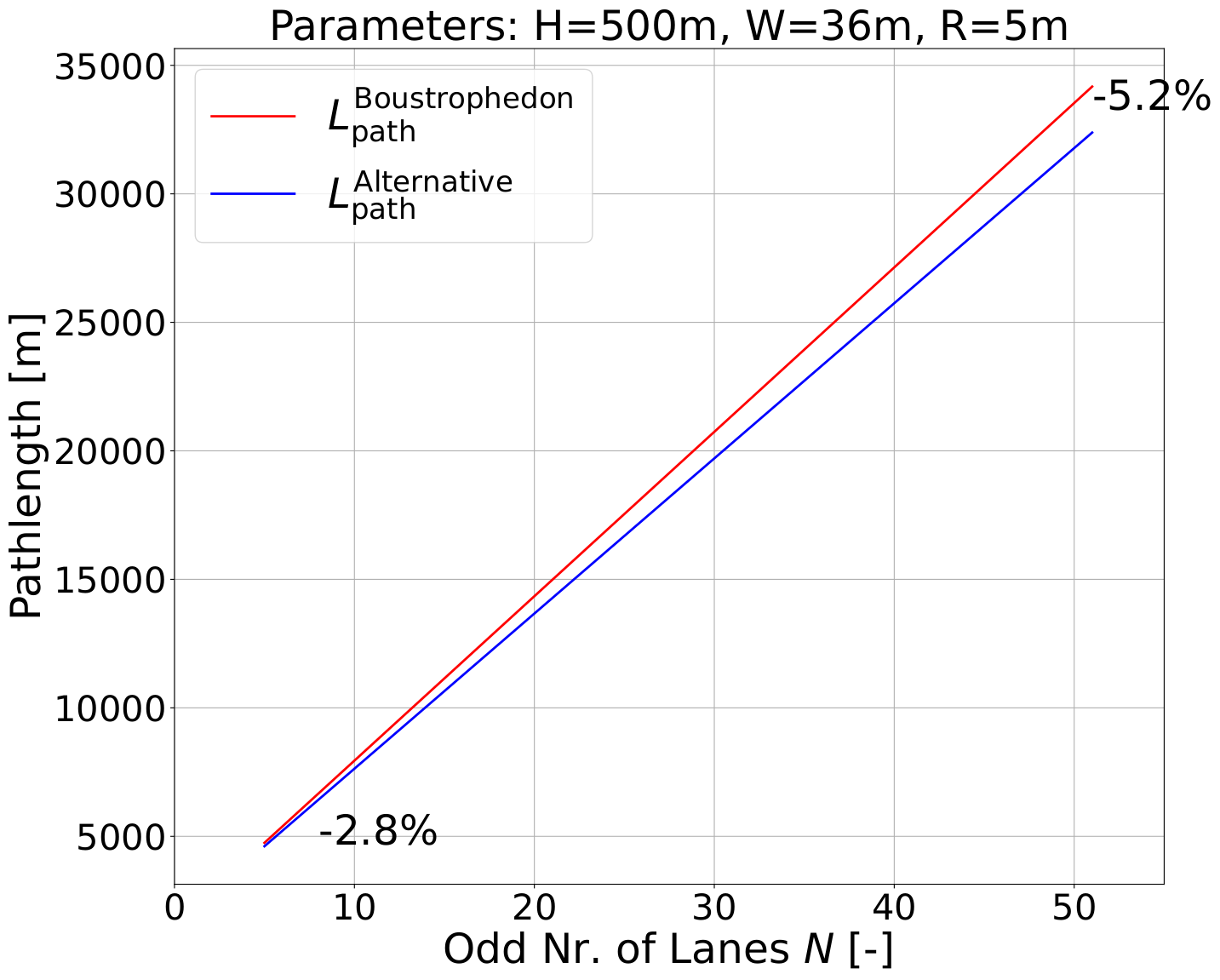}\\[-1pt]
\caption{Pathlengths as a function of the number of mainfield lanes $N>0$ in four scenarios with $W\in\{12\text{m},36\text{m}\}$ and $H\in\{100\text{m},500\text{m}\}$.\\[10pt]}
  \label{fig_oddN_L}
\end{subfigure}
\begin{subfigure}[t]{.9999\linewidth}
  \centering
  \includegraphics[width=.37\linewidth]{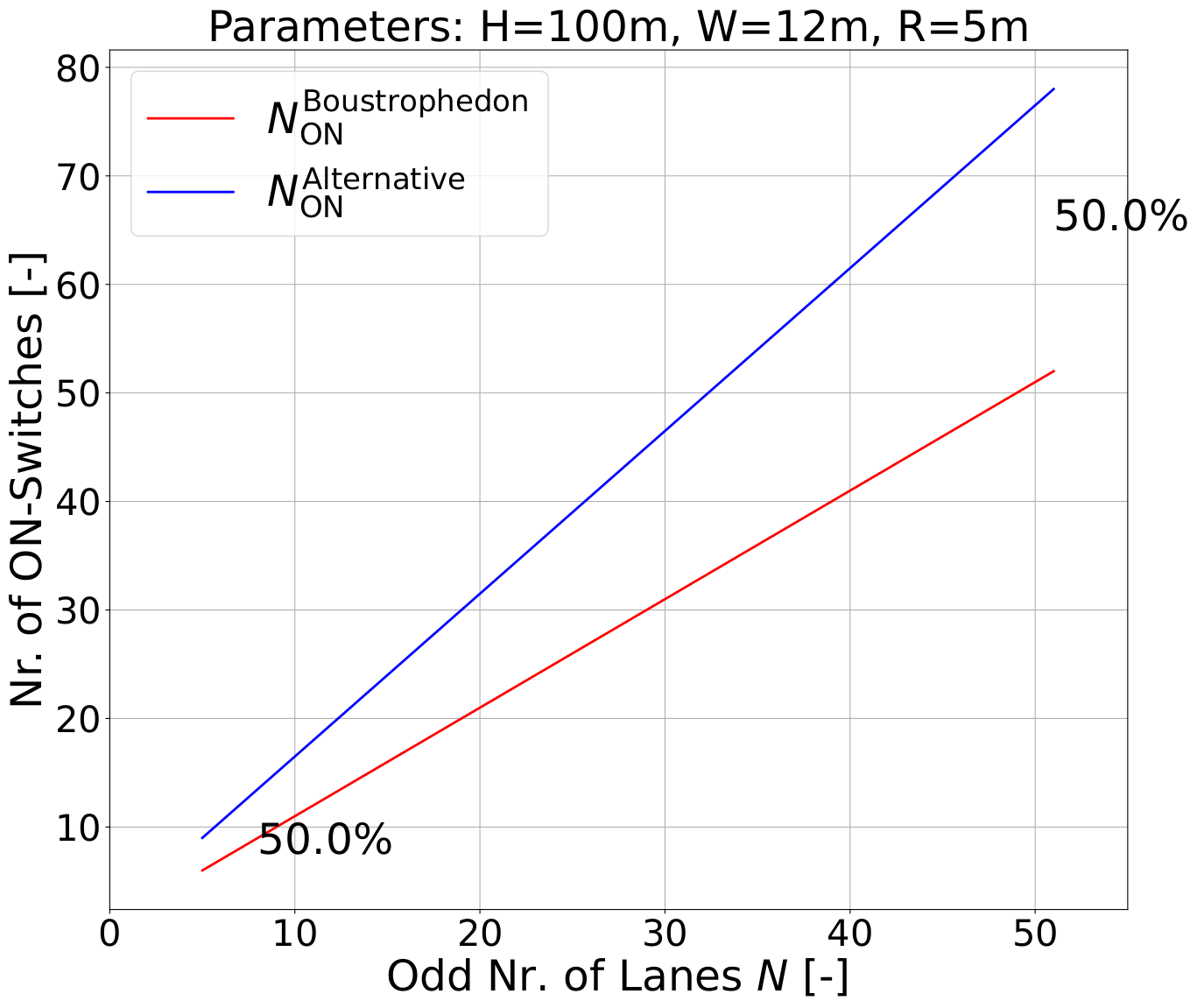}~
  \includegraphics[width=.37\linewidth]{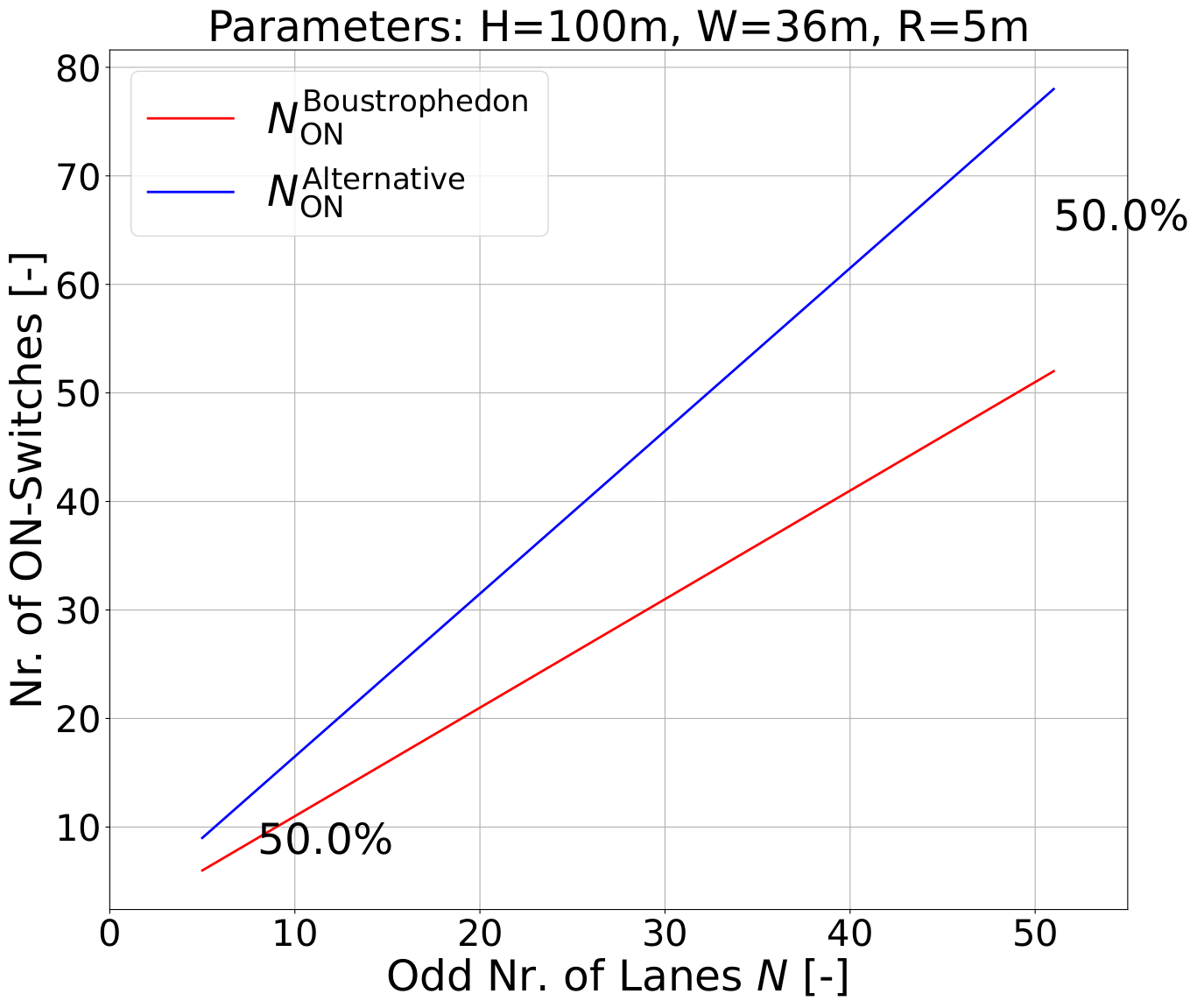}\\[-1pt]
  \includegraphics[width=.37\linewidth]{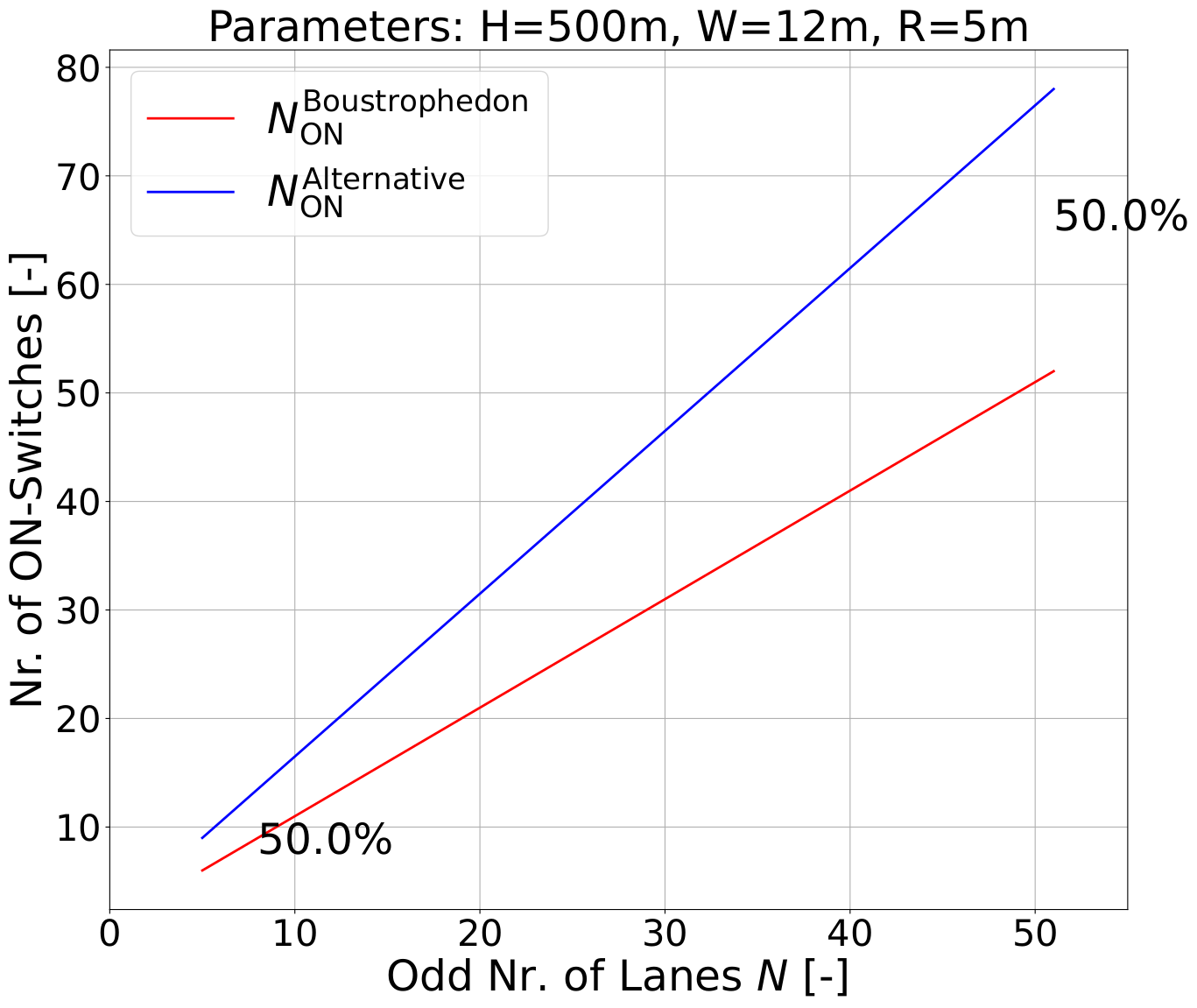}~
  \includegraphics[width=.37\linewidth]{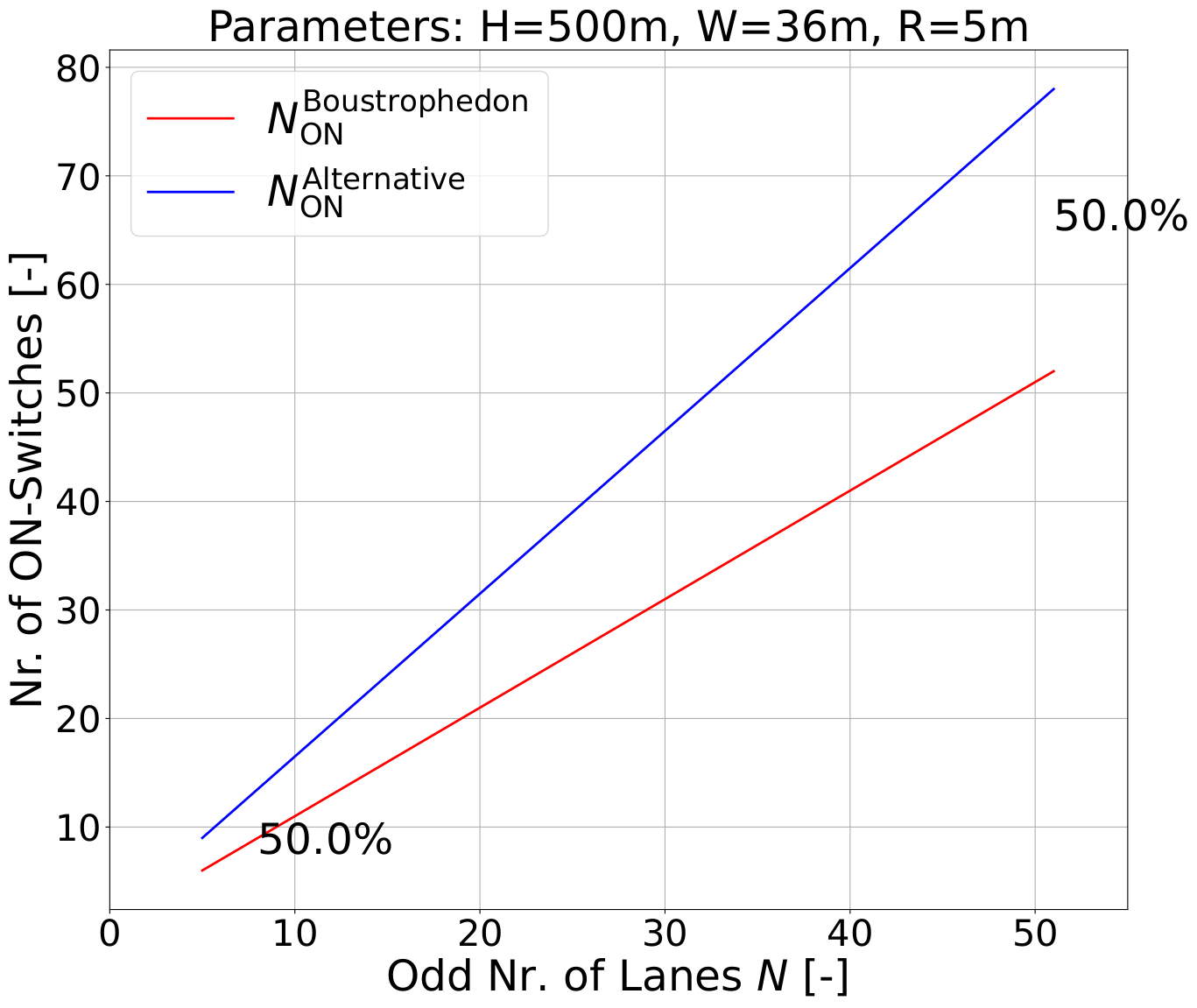}\\[-1pt]
\caption{Number of switching-on states as a function of the number of mainfield lanes $N>0$ in four scenarios with $W\in\{12\text{m},36\text{m}\}$ and $H\in\{100\text{m},500\text{m}\}$.}
  \label{fig_oddN_N}
\end{subfigure}
\caption{Experiments: Visualisation of formulas \eqref{eq_oddN} for an odd number of mainfield lanes $N>0$.}
\label{fig_oddN}
\end{figure*}
%

\begin{figure*}
\vspace*{-1cm}\centering
\begin{subfigure}[t]{.9999\linewidth}
  \centering
  \includegraphics[width=.37\linewidth]{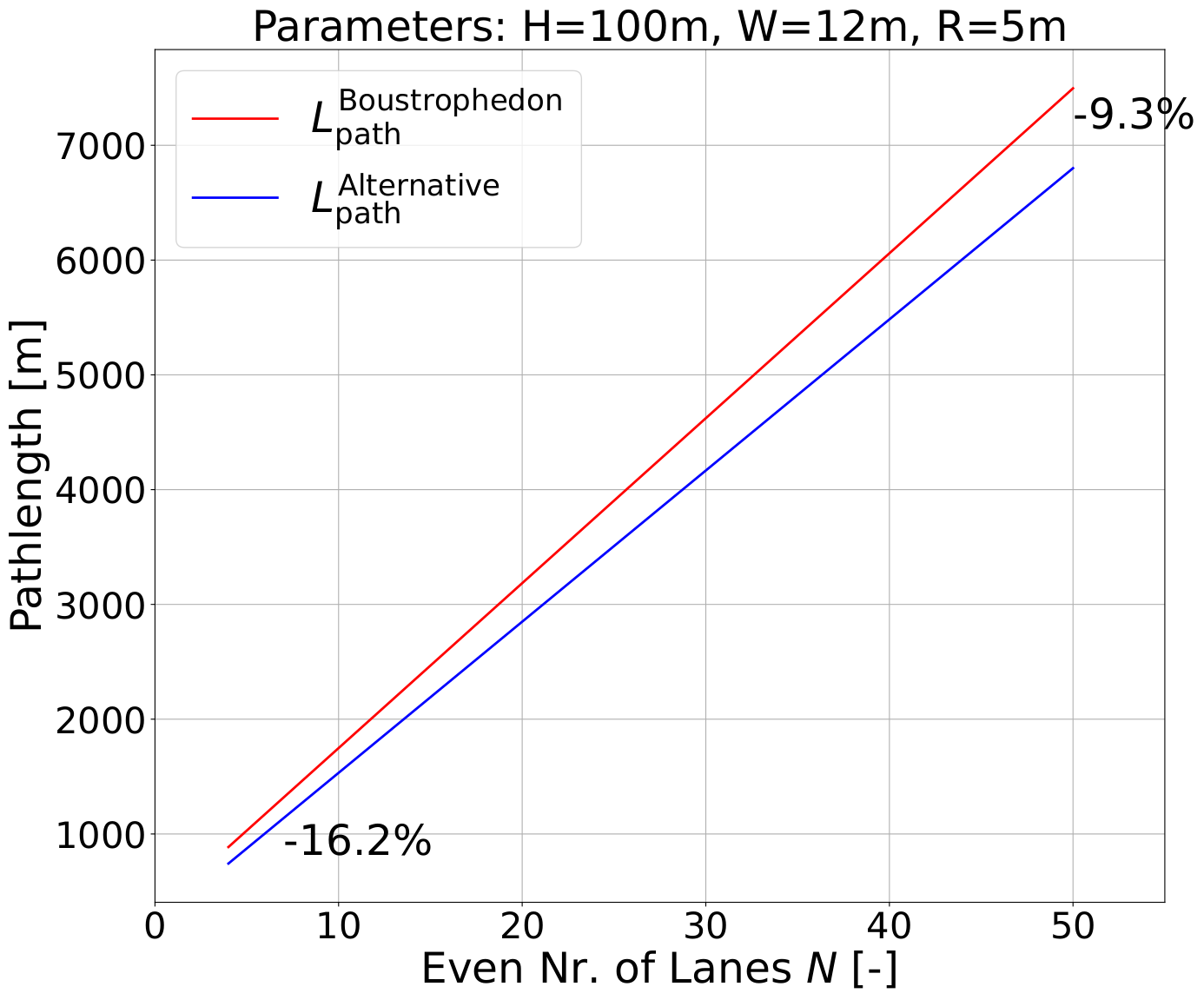}~
  \includegraphics[width=.37\linewidth]{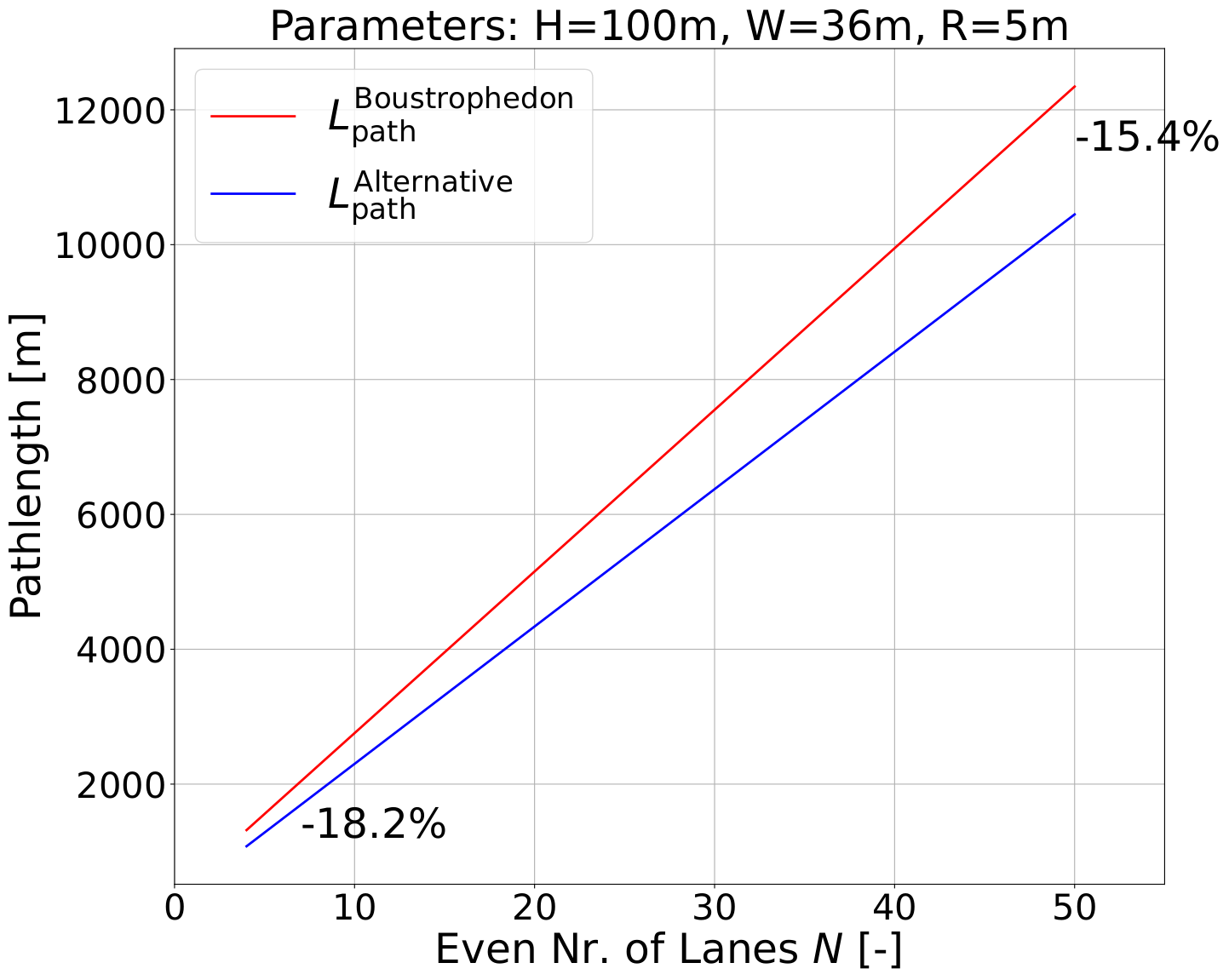}\\[-1pt]
  \includegraphics[width=.37\linewidth]{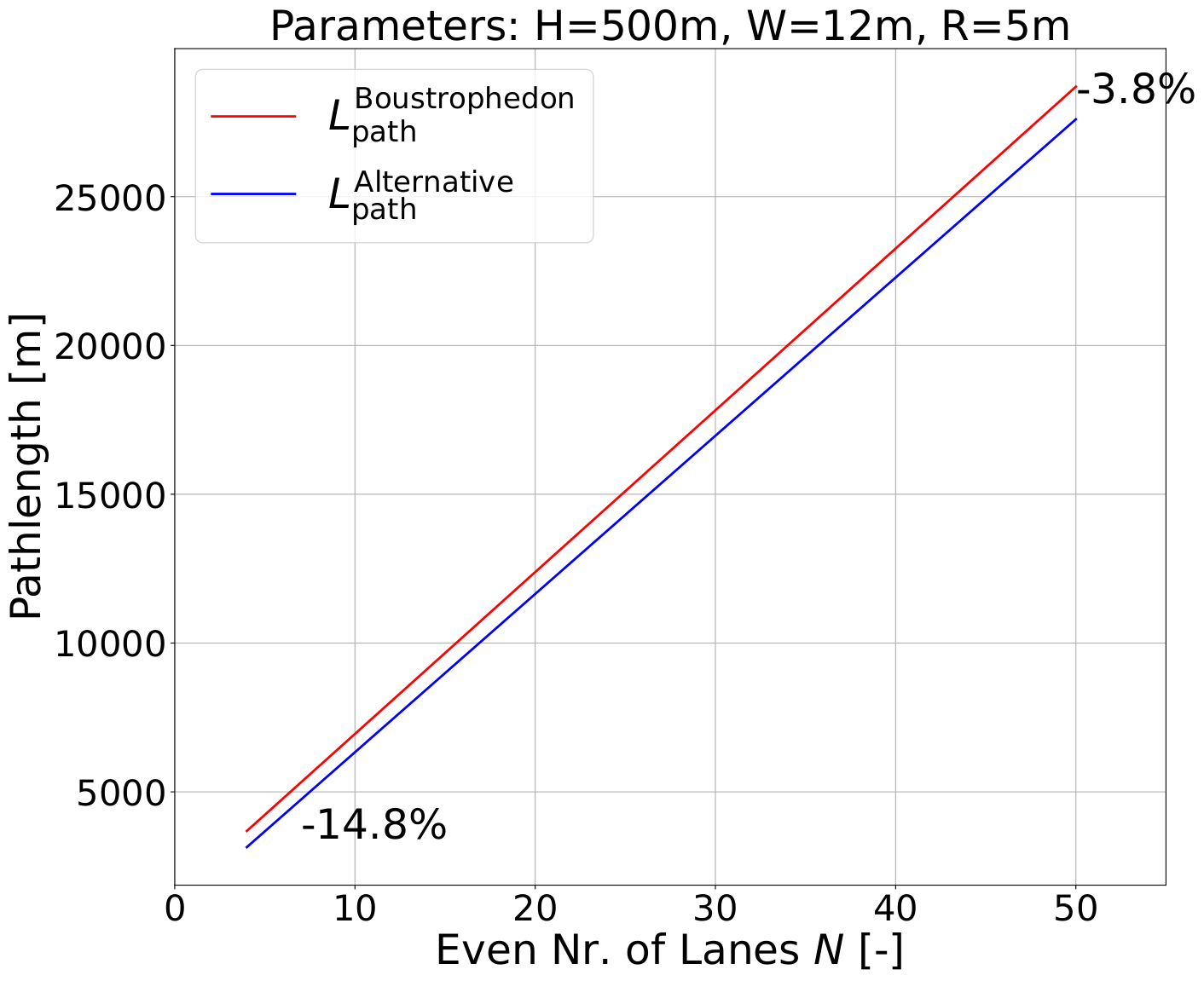}~
  \includegraphics[width=.37\linewidth]{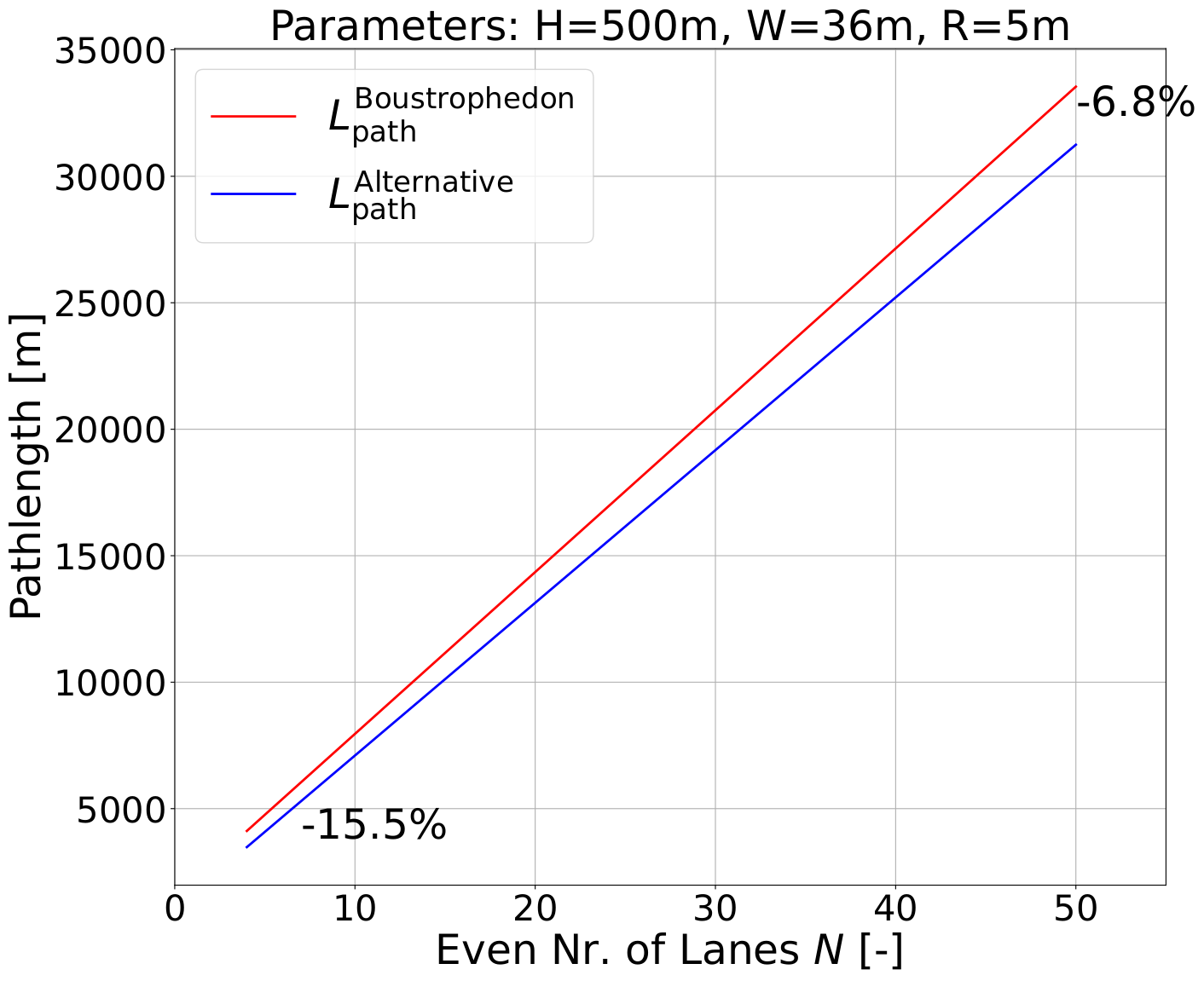}\\[-1pt]
\caption{Pathlengths as a function of the number of mainfield lanes $N>0$ in four scenarios with $W\in\{12\text{m},36\text{m}\}$ and $H\in\{100\text{m},500\text{m}\}$.\\[10pt]}
  \label{fig_evenN_L}1
\end{subfigure}
\begin{subfigure}[t]{.9999\linewidth}
  \centering
  \includegraphics[width=.37\linewidth]{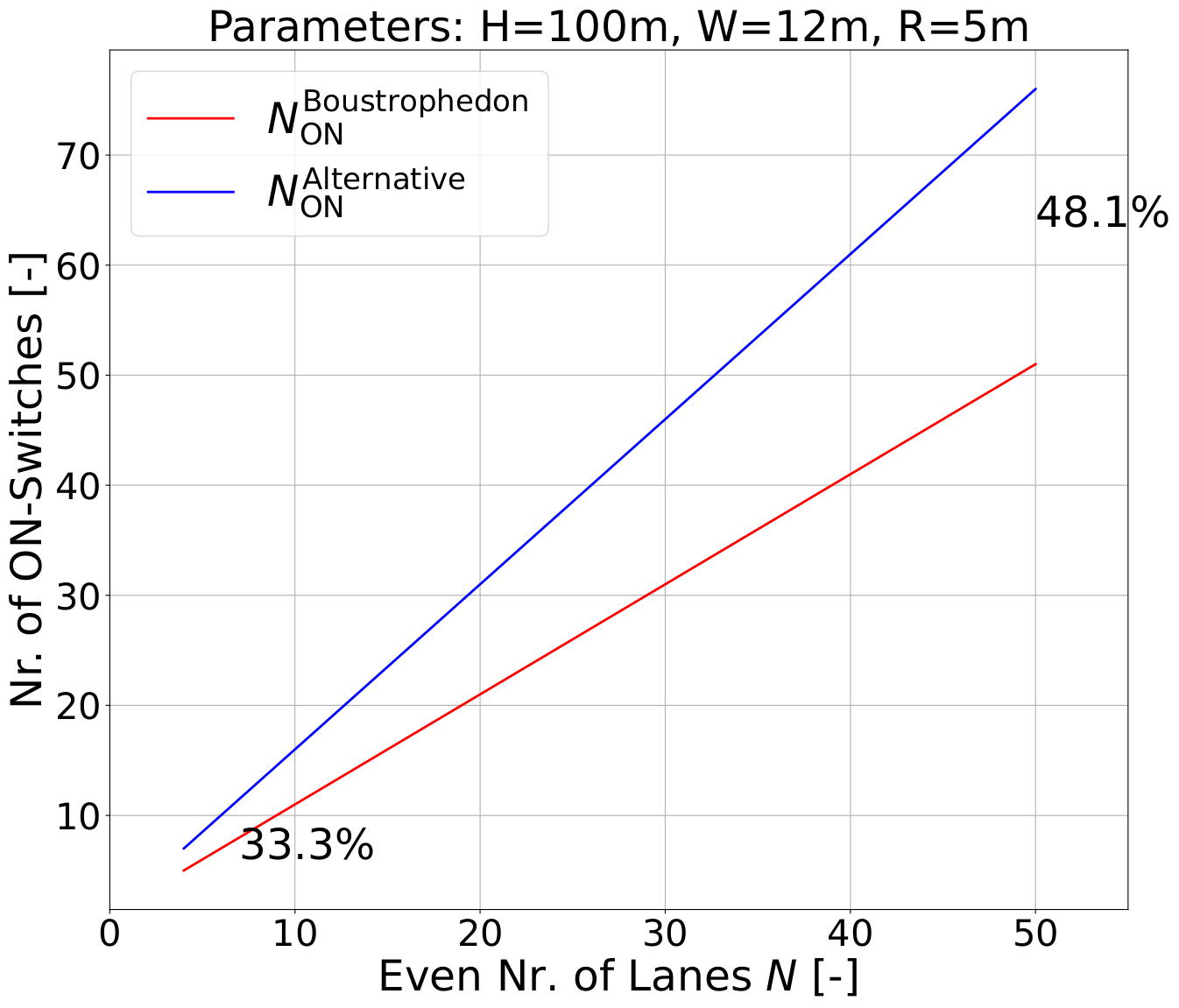}~
  \includegraphics[width=.37\linewidth]{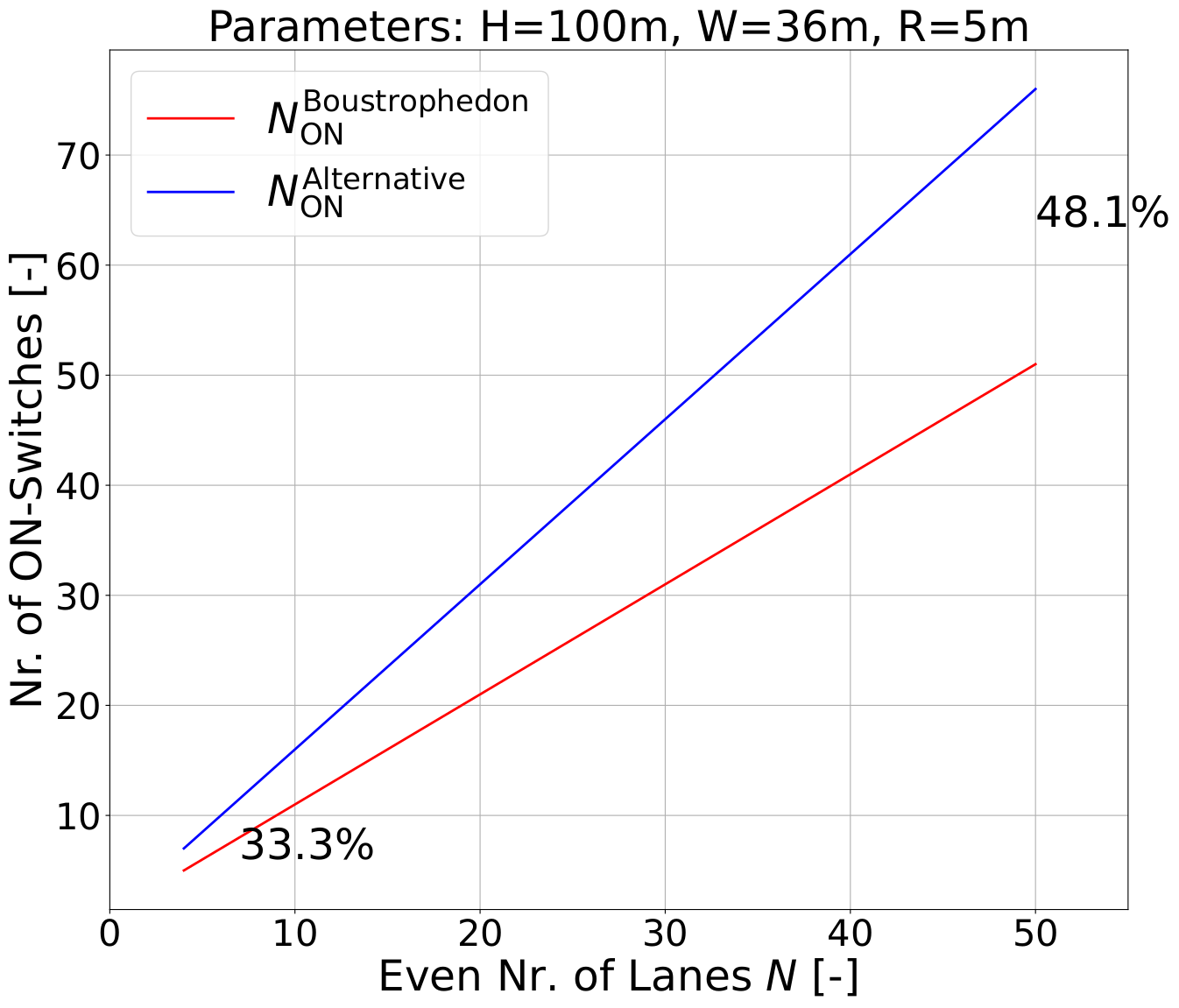}\\[-1pt]
  \includegraphics[width=.37\linewidth]{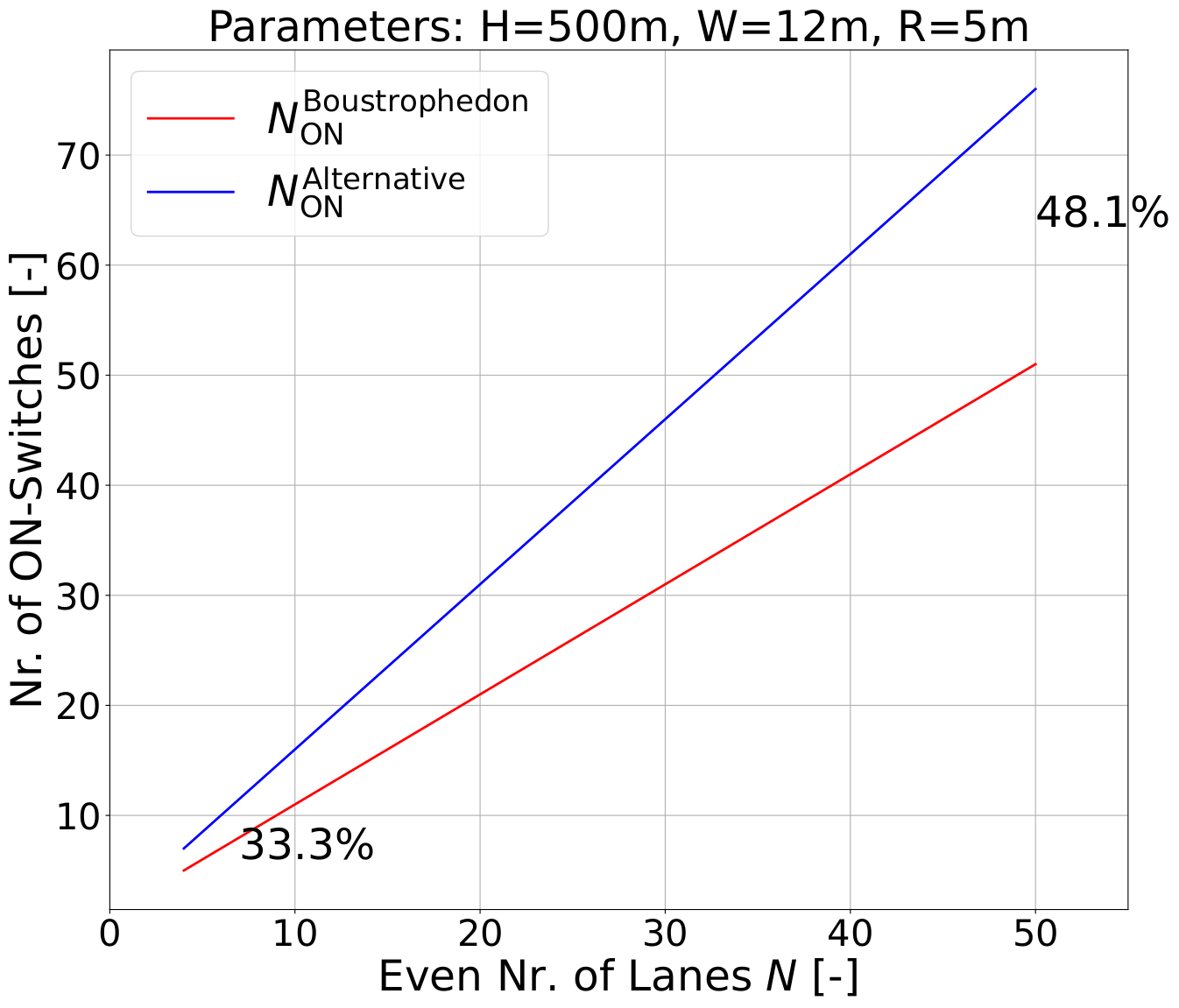}~
  \includegraphics[width=.37\linewidth]{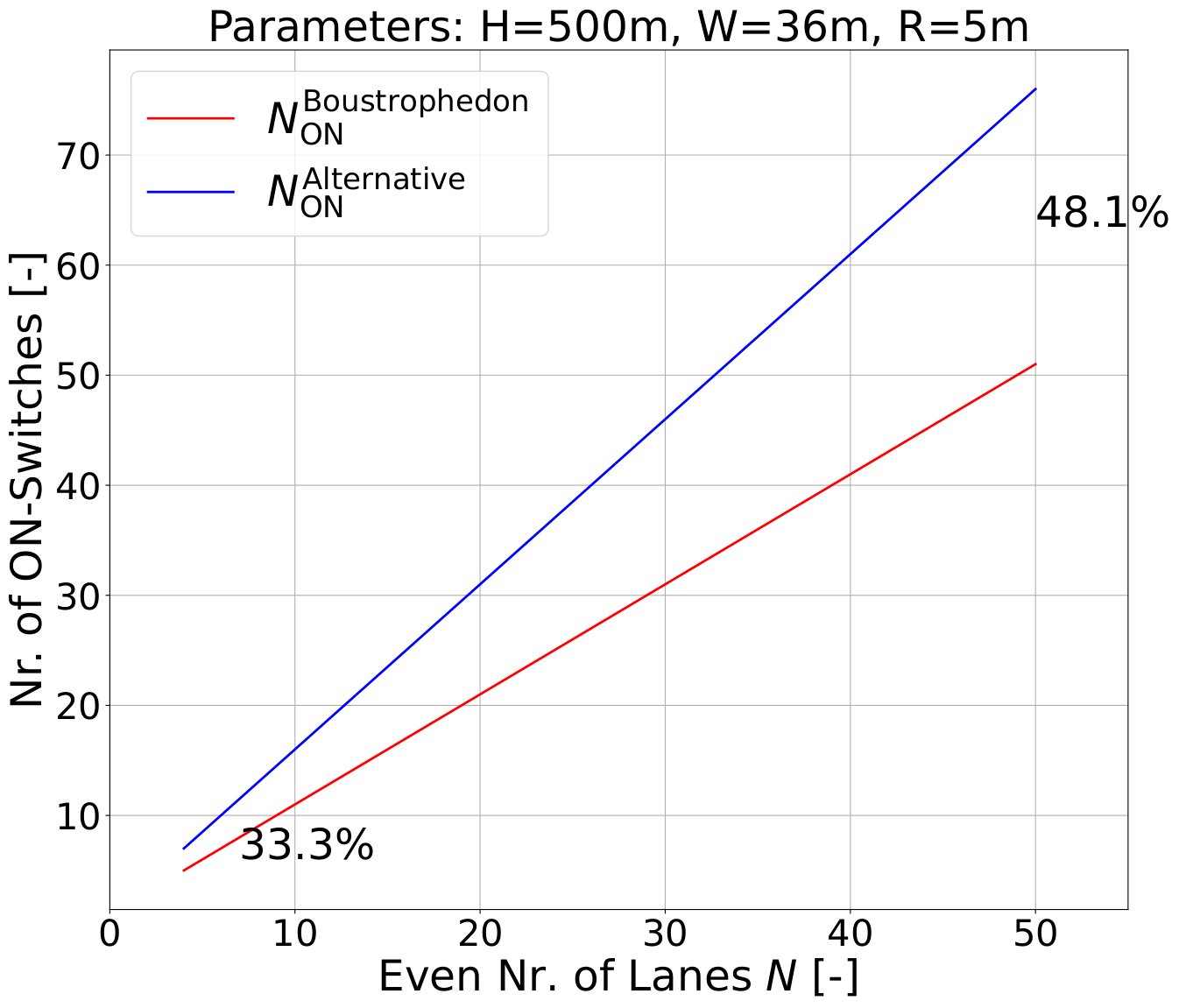}\\[-1pt]
\caption{Number of switching-on states as a function of the number of mainfield lanes $N>0$ in four scenarios with $W\in\{12\text{m},36\text{m}\}$ and $H\in\{100\text{m},500\text{m}\}$.}
  \label{fig_evenN_N}
\end{subfigure}
\caption{Experiments: Visualisation of formulas \eqref{eq_evenN} for an even number of mainfield lanes $N>0$.}
\label{fig_evenN}
\end{figure*}
%

\begin{table}
\vspace{0.3cm}
\centering
\begin{tabular}{|l|r|r|}
\hline 
\rowcolor[gray]{0.8} & & \\[-8pt]
\rowcolor[gray]{0.8} Example Setup & \multicolumn{1}{r|}{Boustrophedon} & \multicolumn{1}{r|}{Alternative} \\[2pt]
\hline 
$W,~H,~\text{Area}$  & $L_\text{path},N_\text{ON}$ & $L_\text{path},~N_\text{ON}$ \\ 
$N$  &  & $\Delta L_\text{path}^\text{m},~\Delta N_\text{ON}$ \\[1pt]
\hline\hline
12m, 100m, 0.7ha & 1020m, 6   & 982m, 9 \\
5  &       & \textbf{-38m}, 3 \\
\hline
12m, 100m, 6.2ha & 7630m, 52   & 7040m, 78 \\
51  &       & \textbf{-590m}, 26 \\
\hline
12m, 500m, 3.6ha & 4220m, 6   & 4182m, 9 \\
5  &       & \textbf{-38m}, 3 \\
\hline
12m, 500m, 31.2ha & 29230m, 52   & 28640m, 78 \\
51  &       & \textbf{-590m}, 26 \\
\hline\hline
36m, 100m, 2.2ha & 1548m, 6   & 1414m, 9 \\
5  &       & \textbf{-134m}, 3 \\
\hline
36m, 100m, 18.7ha & 12574m, 52   & 10784m, 78 \\
51  &       & \textbf{-1790m}, 26 \\
\hline
36m, 500m, 10.8ha & 4748m, 6   & 4614m, 9 \\
5  &       & \textbf{-134m}, 3 \\
\hline
36m, 500m, 93.6ha & 34174m, 52   & 32384m, 78 \\
51  &       & \textbf{-1790m}, 26 \\
\hline
\end{tabular}
\caption{Experiments: Results for an odd number of mainfield lanes $N>0$. A turning radius of $R=5$m is assumed throughout. For formulas and qualitative evaluation see \eqref{eq_oddN} and Fig. \ref{fig_oddN}, respectively.}
\label{tab_oddN}
\end{table}

\begin{table}
\vspace{0.3cm}
\centering
\begin{tabular}{|l|r|r|}
\hline 
\rowcolor[gray]{0.8} & & \\[-8pt]
\rowcolor[gray]{0.8} Example Setup & Boustrophedon & Alternative \\[2pt]
\hline 
$W,~H,~\text{Area}$  & $L_\text{path},~N_\text{ON}$ & $L_\text{path},~N_\text{ON}$ \\ 
$N$  &  & $\Delta L_\text{path}^\text{m},~\Delta N_\text{ON}$ \\[1pt]
\hline\hline
12m, 100m, 0.6ha & 886m, 5   & 742m, 7 \\
4  &       & \textbf{-144m}, 2 \\
\hline
12m, 100m, 6.1ha & 7497m, 51   & 6801m, 76 \\
50  &       & \textbf{-696m}, 25 \\
\hline
12m, 500m, 3.0ha & 3686m, 5   & 3142m, 7 \\
4  &       & \textbf{-544m}, 2 \\
\hline
12m, 500m, 30.6ha & 28697m, 51   & 27601m, 76 \\
50  &       & \textbf{-1096m}, 25 \\
\hline\hline
36m, 100m, 1.8ha & 1318m, 5   & 1078m, 7 \\
4  &       & \textbf{-240m}, 2 \\
\hline
36m, 100m, 18.4ha & 12345m, 51   & 10449m, 76 \\
50  &       & \textbf{-1896m}, 25 \\
\hline
36m, 500m, 9.0ha & 4118m, 5   & 3478m, 7 \\
4  &       & \textbf{-640m}, 2 \\
\hline
36m, 500m, 91.8ha & 33545m, 51   & 31249m, 76 \\
50  &       & \textbf{-2296m}, 25 \\
\hline
\end{tabular}
\caption{Experiments: Results for an even number of mainfield lanes $N>0$. A turning radius of $R=5$m is assumed throughout. For formulas and qualitative evaluation see \eqref{eq_evenN} and Fig. \ref{fig_evenN}, respectively.}
\label{tab_evenN}
\end{table}

For the experimental setup displayed in Fig. \ref{fig_exptsetup} analytical formulas can be derived for the pathlength as well as for the number of required switching-on and switching-off states. For this calculation it has to be distinguished between an odd and an even number of mainfield lanes $N$. 

For an odd number of mainfield lanes, the following formulas can be derived analytically:
\begin{subequations}
\begin{align}
L_\text{path}^\text{Boustrophedon}(N) &= N(H-4R+2\frac{2R\pi}{4} + 4W) + c_1,\label{subeq_L_oddN_Boustro}\\
N_\text{ON}^\text{Boustrophedon}(N) &= N+1,\label{subeq_N_oddN_Boustro}\\
L_\text{path}^\text{Alternative}(N) &= N(H-4R+2\frac{2R\pi}{4} + 3W) + c_2,\label{subeq_L_oddN_Altern}\\
N_\text{ON}^\text{Alternative}(N) &= \frac{3}{2}(N+1),\label{subeq_N_oddN_Altern}
\end{align}
\label{eq_oddN}
\end{subequations}
with offset constants $c_1=(3H-14R+6\frac{2R\pi}{4}+2W)$ and $c_2=(3H-12R+6\frac{2R\pi}{4}+3W)$.

For an even number of mainfield lanes, the following formulas can be derived analytically:
\begin{subequations}
\begin{align}
L_\text{path}^\text{Boustrophedon}(N) &= N(H-4R+2\frac{2R\pi}{4} + 4W) + c_3,\label{subeq_L_evenN_Boustro}\\
N_\text{ON}^\text{Boustrophedon}(N) &= N+1,\label{subeq_N_evenN_Boustro}\\
L_\text{path}^\text{Alternative}(N) &= N(H-4R+2\frac{2R\pi}{4} + 3W) + c_4,\label{subeq_L_evenN_Altern}\\
N_\text{ON}^\text{Alternative}(N) &= \frac{3}{2}N+1,\label{subeq_N_evenN_Altern}
\end{align}
\label{eq_evenN}
\end{subequations}
with offset constants $c_3=(3H-12R+6\frac{2R\pi}{4}+2W)$ and $c_4=(2H-8R+4\frac{2R\pi}{4}+2W)$.

For visualisation, formulas \eqref{eq_oddN} are evaluated and displayed in Fig. \ref{fig_oddN} and Table \ref{tab_oddN}. Similarly, formulas \eqref{eq_evenN} are visualised in Fig. \ref{fig_evenN} and Table \ref{tab_evenN}.

Several comments can be made. First, the key insight is that in both cases the difference in pathlengths, and thus the pathlength savings that can be achieved by the Alternative method, scales linearly with the number of mainfield lanes and the working width,
\begin{equation}
L_\text{path}^\text{Alternative}(N)-L_\text{path}^\text{Boustrophedon}(N)=\begin{cases} -NW + \tilde{c_{21}}, & \text{if}~N~\text{odd}, \\ -NW + \tilde{c_{43}}, & \text{if}~N~\text{even},
\end{cases}\label{eq_diffL}
\end{equation}
where constants are $\tilde{c_{21}}=c_2-c_1$ and $\tilde{c_{43}}=c_4-c_3$. Thus, the more mainfield lanes are needed to cover an agricultural field or work area the more pathlength savings can be achieved in absolute values by the Alternative method. 

Second, while the Alternative method offers pathlength savings as a benefit the Boustrophedon-based method offers savings in the required number of switchings: 
\begin{equation}
N_\text{ON}^\text{Alternative}(N)-N_\text{ON}^\text{Boustrophedon}(N)  =\begin{cases}  \frac{1}{2}(N + 1), & \text{if}~N~\text{odd}, \\ \frac{1}{2}N, & \text{if}~N~\text{even}.
\end{cases}\label{eq_diffN}
\end{equation}

Note that while \eqref{eq_diffL} and \eqref{eq_diffN} were derived for the experimental setup in Fig. \ref{fig_exptsetup}, similar analytical formulas can be derived for alternative starting positions (field entrance) located at a different location along the headland path. This changes the offset constants in \eqref{eq_diffL} and \eqref{eq_diffN}, however, importantly the linear relationships remain the same.

Thus, both the Boustrophedon-based and Alternative method have one advantage and one disadvantage with respect to each other. 

What remains to evaluate is whether pathlength savings or savings in the number of switches are of greater importance. Therefore, in ordert to better get a sense of orders of magnitudes and savings potential, formulas are evaluated for a range of typical parameters. The results are summarised in Table \ref{tab_oddN} and \ref{tab_evenN}.

For example, for an agricultural field of size 18.4ha with an even number of lanes $N=50$, a working width $W=36$m and $H=100$m, the pathlength savings for the Alternative method would be -1896m and 25 more switching-on states in comparison to the Boustrophedon method. Here, several comments can be made.

First, the larger number of required switching-on states is undesired. This is since each switching generates a real-world transient behaviour (without aforementioned nominal assumption of instant spray application) during which only partial spray is applied while the nozzle switching occurs. Thus, spray uniformity is lost during transient distances. To put this in perspective, suppose this switching transient occurs for 2m (e.g. a traveling speed of 7.2km/h and a switching transient duration of 1s) for each switching while the machinery is traveling in the agricultural field. Then, 25 more switching-on states and their corresponding switching-off states combinedly imply 100m pathlength of the total pathlength where transient switching behavior occurs. Note that these 100m stand in contrast to -1896m pathlength savings that can be achieved through the Alternative method. This discussion is given to underline different orders of magnitude.


Pathlength savings imply (i) time savings and (ii) fuel savings for the machinery. For example, pathlength savings of -1896m at a traveling speed of 5km/h imply 22.8min time savings. This is significant. At a traveling speed of 10km/h the time savings are still 11.4min.

Whether to more highly value pathlength savings or instead to prefer fewer nozzle switchings is a decision any practitioner has to decide for themselves. On the one hand there are time savings and fuel savings, on the other hand there is the avoidance of switching transients where spray is applied only partially. From an economic point of view time and fuel savings are arguably more important.

Finally, a detail about the effect of an odd or even number of mainfield lanes on pathlength savings is discussed. As Fig. \ref{fig_oddN_L} illustrates for a small odd number of mainfield lanes, $N=5$, the pathlength savings of the Alternative method with respect to the Boustrophedon-based method are between $-0.9\%$ and $-8.7\%$. In contrast, as Fig. \ref{fig_evenN_L} illustrates for a small even number of mainfield lanes, $N=4$, the pathlength savings are between $-14.8\%$ and $-18.2\%$. The reason for this large discrepancy in pathlength savings is which one of the two cases in Fig. \ref{fig_2specialcases} applies. For the given examples, Fig. \ref{fig_2specialcases_1} applies for an even number for $N$, whereas Fig. \ref{fig_2specialcases_2} applies for an odd number for $N$. Thus, for an odd number $N$ the right-most headland path segment is traversed \emph{twice} according to Fig. \ref{fig_2specialcases_2}. This is unavoidable and necessary to cover the penultimate mainfield lane, however, results in smaller pathlength savings with respect to the Boustrophedon-based method. The smaller $N$ and the larger the mainfield length $H$, the more profound this loss of pathlength savings potential is. In contrast, for an \emph{even} number of mainfield lanes maximal pathlength savings potential is achieved by the Alternative method with respect to the Boustrophedon-based method.

\section{Conclusion\label{sec_conclusion}}

This paper presented a predictive logic for the on- and off-switching of a set of nozzles attached to a boom aligned along a working width for a specific path pattern for area coverage. The path pattern is efficient for area coverage in that its concatenation yields shorter area coverage pathlengths than an alternative Boustrophedon-based area coverage path. The proposed switching logic for the path pattern is efficient in that it avoids switching-on states during turn maneuvers by exploiting the special structure of the path pattern. 

Two predictive aspects of the proposed switching logic were highlighted, first within its framework for one path pattern and then within the framework of concatenating multiple path patterns.

The method was compared to a state-of-the-art reactive switching logic for Boustrophedon-based area coverage path planning.

Assuming a convexly shaped work area, one advantage and one disadvantage of proposed method were highlighted. The advantage is pathlength savings that scale linearly with the number of mainfield lanes and the working width. The disadvantage is that the number of required switching-on states is larger than for the Boustrophedon-based method and scales linearly with 50\% times the closest rounded up even number of mainfield lanes.

The implications of pathlength savings for time and fuel savings were hinted and numerical examples for illustration were given. The implications of short transients during switching changes, in which only partial spray is applied,  was discussed.

Future work will analyse the effect of proposed switching logic for non-convexly shaped work areas.

\bibliographystyle{model5-names} 
\bibliography{mybibfile.bib}

\begin{thebibliography}{30}
\expandafter\ifx\csname natexlab\endcsname\relax\def\natexlab#1{#1}\fi
\providecommand{\url}[1]{\texttt{#1}}
\providecommand{\href}[2]{#2}
\providecommand{\path}[1]{#1}
\providecommand{\DOIprefix}{doi:}
\providecommand{\ArXivprefix}{arXiv:}
\providecommand{\URLprefix}{URL: }
\providecommand{\Pubmedprefix}{pmid:}
\providecommand{\doi}[1]{\href{http://dx.doi.org/#1}{\path{#1}}}
\providecommand{\Pubmed}[1]{\href{pmid:#1}{\path{#1}}}
\providecommand{\bibinfo}[2]{#2}
\ifx\xfnm\relax \def\xfnm[#1]{\unskip,\space#1}\fi
\bibitem[{Al-Mallahi et~al.(2023)Al-Mallahi, Natarajan \&
  Shirzadifar}]{al2023development}
\bibinfo{author}{Al-Mallahi, A.}, \bibinfo{author}{Natarajan, M.}, \&
  \bibinfo{author}{Shirzadifar, A.} (\bibinfo{year}{2023}).
\newblock \bibinfo{title}{Development of robust communication algorithm between
  machine vision and boom sprayer for spot application via iso 11783}.
\newblock {\it \bibinfo{journal}{Smart Agricultural Technology}\/},  {\it
  \bibinfo{volume}{4}\/}, \bibinfo{pages}{100212}.
\bibitem[{Asiminari et~al.(2024)Asiminari, Moysiadis, Kateris, Busato, Wu,
  Achillas, S{\o}rensen, Pearson \& Bochtis}]{asiminari2024integrated}
\bibinfo{author}{Asiminari, G.}, \bibinfo{author}{Moysiadis, V.},
  \bibinfo{author}{Kateris, D.}, \bibinfo{author}{Busato, P.},
  \bibinfo{author}{Wu, C.}, \bibinfo{author}{Achillas, C.},
  \bibinfo{author}{S{\o}rensen, C.~G.}, \bibinfo{author}{Pearson, S.}, \&
  \bibinfo{author}{Bochtis, D.} (\bibinfo{year}{2024}).
\newblock \bibinfo{title}{Integrated route-planning system for agricultural
  robots}.
\newblock {\it \bibinfo{journal}{AgriEngineering}\/},  {\it
  \bibinfo{volume}{6}\/}, \bibinfo{pages}{657--677}.
\bibitem[{Burgers et~al.(2021)Burgers, Gaard \&
  Hyronimus}]{burgers2021comparison}
\bibinfo{author}{Burgers, T.~A.}, \bibinfo{author}{Gaard, J.~D.}, \&
  \bibinfo{author}{Hyronimus, B.~J.} (\bibinfo{year}{2021}).
\newblock \bibinfo{title}{Comparison of three commercial automatic boom height
  systems for agricultural sprayers}.
\newblock {\it \bibinfo{journal}{Applied Engineering in Agriculture}\/},  {\it
  \bibinfo{volume}{37}\/}, \bibinfo{pages}{287--298}.
\bibitem[{Carre{\~n}o~Ruiz et~al.(2022)Carre{\~n}o~Ruiz, Bloise, Guglieri \&
  D’Ambrosio}]{carreno2022numerical}
\bibinfo{author}{Carre{\~n}o~Ruiz, M.}, \bibinfo{author}{Bloise, N.},
  \bibinfo{author}{Guglieri, G.}, \& \bibinfo{author}{D’Ambrosio, D.}
  (\bibinfo{year}{2022}).
\newblock \bibinfo{title}{Numerical analysis and wind tunnel validation of
  droplet distribution in the wake of an unmanned aerial spraying system in
  forward flight}.
\newblock {\it \bibinfo{journal}{Drones}\/},  {\it \bibinfo{volume}{6}\/},
  \bibinfo{pages}{329}.
\bibitem[{Galceran \& Carreras(2013)}]{galceran2013survey}
\bibinfo{author}{Galceran, E.}, \& \bibinfo{author}{Carreras, M.}
  (\bibinfo{year}{2013}).
\newblock \bibinfo{title}{A survey on coverage path planning for robotics}.
\newblock {\it \bibinfo{journal}{Robotics and Autonomous systems}\/},  {\it
  \bibinfo{volume}{61}\/}, \bibinfo{pages}{1258--1276}.
\bibitem[{Hassen et~al.(2013)Hassen, Sidik \& Sheriff}]{hassen2013effect}
\bibinfo{author}{Hassen, N.~S.}, \bibinfo{author}{Sidik, N. A.~C.}, \&
  \bibinfo{author}{Sheriff, J.~M.} (\bibinfo{year}{2013}).
\newblock \bibinfo{title}{Effect of nozzle type, angle and pressure on spray
  volumetric distribution of broadcasting and banding application}.
\newblock {\it \bibinfo{journal}{Journal of Mechanical Engineering
  Research}\/},  {\it \bibinfo{volume}{5}\/}, \bibinfo{pages}{76--81}.
\bibitem[{He et~al.(2023)He, Bao, Yu, Lu, He \& Liu}]{he2023dynamic}
\bibinfo{author}{He, Z.}, \bibinfo{author}{Bao, Y.}, \bibinfo{author}{Yu, Q.},
  \bibinfo{author}{Lu, P.}, \bibinfo{author}{He, Y.}, \& \bibinfo{author}{Liu,
  Y.} (\bibinfo{year}{2023}).
\newblock \bibinfo{title}{Dynamic path planning method for headland turning of
  unmanned agricultural vehicles}.
\newblock {\it \bibinfo{journal}{Computers and Electronics in Agriculture}\/},
  {\it \bibinfo{volume}{206}\/}, \bibinfo{pages}{107699}.
\bibitem[{H{\"o}ffmann et~al.(2023)H{\"o}ffmann, Patel \&
  B{\"u}skens}]{hoffmann2023optimal}
\bibinfo{author}{H{\"o}ffmann, M.}, \bibinfo{author}{Patel, S.}, \&
  \bibinfo{author}{B{\"u}skens, C.} (\bibinfo{year}{2023}).
\newblock \bibinfo{title}{Optimal coverage path planning for agricultural
  vehicles with curvature constraints}.
\newblock {\it \bibinfo{journal}{Agriculture}\/},  {\it
  \bibinfo{volume}{13}\/}, \bibinfo{pages}{2112}.
\bibitem[{H{\"o}ffmann et~al.(2024)H{\"o}ffmann, Patel \&
  B{\"u}skens}]{hoffmann2024optimal}
\bibinfo{author}{H{\"o}ffmann, M.}, \bibinfo{author}{Patel, S.}, \&
  \bibinfo{author}{B{\"u}skens, C.} (\bibinfo{year}{2024}).
\newblock \bibinfo{title}{Optimal guidance track generation for precision
  agriculture: A review of coverage path planning techniques}.
\newblock {\it \bibinfo{journal}{Journal of Field Robotics}\/},  {\it
  \bibinfo{volume}{41}\/}, \bibinfo{pages}{823--844}.
\bibitem[{Holterman et~al.(1997)Holterman, Van De~Zande, Porskamp \&
  Huijsmans}]{holterman1997modelling}
\bibinfo{author}{Holterman, H.}, \bibinfo{author}{Van De~Zande, J.},
  \bibinfo{author}{Porskamp, H.}, \& \bibinfo{author}{Huijsmans, J.}
  (\bibinfo{year}{1997}).
\newblock \bibinfo{title}{Modelling spray drift from boom sprayers}.
\newblock {\it \bibinfo{journal}{Computers and electronics in agriculture}\/},
  {\it \bibinfo{volume}{19}\/}, \bibinfo{pages}{1--22}.
\bibitem[{Jayalakshmi et~al.(2025)Jayalakshmi, Nair \&
  Sathish}]{jayalakshmi2025comprehensive}
\bibinfo{author}{Jayalakshmi, K.}, \bibinfo{author}{Nair, V.~G.}, \&
  \bibinfo{author}{Sathish, D.} (\bibinfo{year}{2025}).
\newblock \bibinfo{title}{A comprehensive survey on coverage path planning for
  mobile robots in dynamic environments}.
\newblock {\it \bibinfo{journal}{IEEE Access}\/}, .
\bibitem[{Li et~al.(2023)Li, Nie, Chen, Ge \& Li}]{li2023development}
\bibinfo{author}{Li, J.}, \bibinfo{author}{Nie, Z.}, \bibinfo{author}{Chen,
  Y.}, \bibinfo{author}{Ge, D.}, \& \bibinfo{author}{Li, M.}
  (\bibinfo{year}{2023}).
\newblock \bibinfo{title}{Development of boom posture adjustment and control
  system for wide spray boom}.
\newblock {\it \bibinfo{journal}{Agriculture}\/},  {\it
  \bibinfo{volume}{13}\/}, \bibinfo{pages}{2162}.
\bibitem[{Luck et~al.(2010)Luck, Pitla, Shearer, Mueller, Dillon, Fulton \&
  Higgins}]{luck2010potential}
\bibinfo{author}{Luck, J.}, \bibinfo{author}{Pitla, S.},
  \bibinfo{author}{Shearer, S.}, \bibinfo{author}{Mueller, T.},
  \bibinfo{author}{Dillon, C.}, \bibinfo{author}{Fulton, J.}, \&
  \bibinfo{author}{Higgins, S.} (\bibinfo{year}{2010}).
\newblock \bibinfo{title}{Potential for pesticide and nutrient savings via
  map-based automatic boom section control of spray nozzles}.
\newblock {\it \bibinfo{journal}{Computers and Electronics in Agriculture}\/},
  {\it \bibinfo{volume}{70}\/}, \bibinfo{pages}{19--26}.
\bibitem[{Mangus et~al.(2017)Mangus, Sharda, Engelhardt, Flippo, Strasser, Luck
  \& Griffin}]{mangus2017analyzing}
\bibinfo{author}{Mangus, D.~L.}, \bibinfo{author}{Sharda, A.},
  \bibinfo{author}{Engelhardt, A.}, \bibinfo{author}{Flippo, D.},
  \bibinfo{author}{Strasser, R.}, \bibinfo{author}{Luck, J.~D.}, \&
  \bibinfo{author}{Griffin, T.} (\bibinfo{year}{2017}).
\newblock \bibinfo{title}{Analyzing the nozzle spray fan pattern of an
  agricultural sprayer using pulse width modulation technology to generate an
  on-ground coverage map}.
\newblock {\it \bibinfo{journal}{Transactions of the ASABE}\/},  {\it
  \bibinfo{volume}{60}\/}, \bibinfo{pages}{315--325}.
\bibitem[{Plessen(2025)}]{plessen2025path}
\bibinfo{author}{Plessen, M.} (\bibinfo{year}{2025}).
\newblock \bibinfo{title}{Path planning for spot spraying with uavs combining
  tsp and area coverages}.
\newblock {\it \bibinfo{journal}{Smart Agricultural Technology}\/},  (p.
  \bibinfo{pages}{100965}).
\bibitem[{Plessen(2019)}]{plessen2019optimal}
\bibinfo{author}{Plessen, M.~G.} (\bibinfo{year}{2019}).
\newblock \bibinfo{title}{Optimal in-field routing for full and partial field
  coverage with arbitrary non-convex fields and multiple obstacle areas}.
\newblock {\it \bibinfo{journal}{Biosystems engineering}\/},  {\it
  \bibinfo{volume}{186}\/}, \bibinfo{pages}{234--245}.
\bibitem[{Plessen(2021)}]{plessen2021freeform}
\bibinfo{author}{Plessen, M.~G.} (\bibinfo{year}{2021}).
\newblock \bibinfo{title}{Freeform path fitting for the minimisation of the
  number of transitions between headland path and interior lanes within
  agricultural fields}.
\newblock {\it \bibinfo{journal}{Artificial Intelligence in Agriculture}\/},
  {\it \bibinfo{volume}{5}\/}, \bibinfo{pages}{233--239}.
\bibitem[{Plessen(2018)}]{plessen2018partial}
\bibinfo{author}{Plessen, M. M.~G.} (\bibinfo{year}{2018}).
\newblock \bibinfo{title}{Partial field coverage based on two path planning
  patterns}.
\newblock {\it \bibinfo{journal}{Biosystems engineering}\/},  {\it
  \bibinfo{volume}{171}\/}, \bibinfo{pages}{16--29}.
\bibitem[{Portman(1979)}]{portman1979calibrating}
\bibinfo{author}{Portman, R.~W.} (\bibinfo{year}{1979}).
\newblock \bibinfo{title}{Calibrating single nozzle and boom sprayers}, .
\bibitem[{Pour~Arab et~al.(2023)Pour~Arab, Spisser \&
  Essert}]{pour2023complete}
\bibinfo{author}{Pour~Arab, D.}, \bibinfo{author}{Spisser, M.}, \&
  \bibinfo{author}{Essert, C.} (\bibinfo{year}{2023}).
\newblock \bibinfo{title}{Complete coverage path planning for wheeled
  agricultural robots}.
\newblock {\it \bibinfo{journal}{Journal of Field Robotics}\/},  {\it
  \bibinfo{volume}{40}\/}, \bibinfo{pages}{1460--1503}.
\bibitem[{Saleem et~al.(2023)Saleem, Zaman, Schumann \&
  Naqvi}]{saleem2023variable}
\bibinfo{author}{Saleem, S.~R.}, \bibinfo{author}{Zaman, Q.~U.},
  \bibinfo{author}{Schumann, A.~W.}, \& \bibinfo{author}{Naqvi, S. M. Z.~A.}
  (\bibinfo{year}{2023}).
\newblock \bibinfo{title}{Variable rate technologies: development, adaptation,
  and opportunities in agriculture}.
\newblock In {\it \bibinfo{booktitle}{Precision agriculture}\/} (pp.
  \bibinfo{pages}{103--122}).
\newblock \bibinfo{publisher}{Elsevier}.
\bibitem[{Sequeira(1979)}]{sequeira1979structured}
\bibinfo{author}{Sequeira, E.~G.} (\bibinfo{year}{1979}).
\newblock \bibinfo{title}{Structured vs. unstructured scan path in static
  visual search performance}, .
\bibitem[{Sharda(2011)}]{sharda2011boom}
\bibinfo{author}{Sharda, A.} (\bibinfo{year}{2011}).
\newblock {\it \bibinfo{title}{Boom Dynamics during Control System Response on
  Agricultural Sprayers}\/}.
\newblock Ph.D. thesis Auburn University.
\bibitem[{Smith et~al.(2000)Smith, Oakley, Williams \&
  Kirkpatrick}]{smith2000broadcast}
\bibinfo{author}{Smith, D.}, \bibinfo{author}{Oakley, D.},
  \bibinfo{author}{Williams, E.}, \& \bibinfo{author}{Kirkpatrick, A.}
  (\bibinfo{year}{2000}).
\newblock \bibinfo{title}{Broadcast spray deposits from fan nozzles}.
\newblock {\it \bibinfo{journal}{Applied Engineering in Agriculture}\/},  {\it
  \bibinfo{volume}{16}\/}, \bibinfo{pages}{109--113}.
\bibitem[{Tan et~al.(2021)Tan, Mohd-Mokhtar \& Arshad}]{tan2021comprehensive}
\bibinfo{author}{Tan, C.~S.}, \bibinfo{author}{Mohd-Mokhtar, R.}, \&
  \bibinfo{author}{Arshad, M.~R.} (\bibinfo{year}{2021}).
\newblock \bibinfo{title}{A comprehensive review of coverage path planning in
  robotics using classical and heuristic algorithms}.
\newblock {\it \bibinfo{journal}{IEEE Access}\/},  {\it \bibinfo{volume}{9}\/},
  \bibinfo{pages}{119310--119342}.
\bibitem[{Tao \& Luh(1989)}]{tao1989coordination}
\bibinfo{author}{Tao, J.~M.}, \& \bibinfo{author}{Luh, J.}
  (\bibinfo{year}{1989}).
\newblock \bibinfo{title}{Coordination of two redundant robots}.
\newblock In {\it \bibinfo{booktitle}{1989 IEEE International Conference on
  Robotics and Automation}\/} (pp. \bibinfo{pages}{425--426}).
\newblock \bibinfo{organization}{IEEE Computer Society}.
\bibitem[{Utamima \& Djunaidy(2022)}]{utamima2022agricultural}
\bibinfo{author}{Utamima, A.}, \& \bibinfo{author}{Djunaidy, A.}
  (\bibinfo{year}{2022}).
\newblock \bibinfo{title}{Agricultural routing planning: A narrative review of
  literature}.
\newblock {\it \bibinfo{journal}{Procedia Computer Science}\/},  {\it
  \bibinfo{volume}{197}\/}, \bibinfo{pages}{693--700}.
\bibitem[{Vijayakumar et~al.(2023)Vijayakumar, Ampatzidis, Schueller \&
  Burks}]{vijayakumar2023smart}
\bibinfo{author}{Vijayakumar, V.}, \bibinfo{author}{Ampatzidis, Y.},
  \bibinfo{author}{Schueller, J.~K.}, \& \bibinfo{author}{Burks, T.}
  (\bibinfo{year}{2023}).
\newblock \bibinfo{title}{Smart spraying technologies for precision weed
  management: A review}.
\newblock {\it \bibinfo{journal}{Smart Agricultural Technology}\/},  {\it
  \bibinfo{volume}{6}\/}, \bibinfo{pages}{100337}.
\bibitem[{Wang et~al.(2023{\natexlab{a}})Wang, Zhang, Song, Yu, Shan, Gu \&
  Lan}]{wang2023evaluation}
\bibinfo{author}{Wang, G.}, \bibinfo{author}{Zhang, T.}, \bibinfo{author}{Song,
  C.}, \bibinfo{author}{Yu, X.}, \bibinfo{author}{Shan, C.},
  \bibinfo{author}{Gu, H.}, \& \bibinfo{author}{Lan, Y.}
  (\bibinfo{year}{2023}{\natexlab{a}}).
\newblock \bibinfo{title}{Evaluation of spray drift of plant protection drone
  nozzles based on wind tunnel test}.
\newblock {\it \bibinfo{journal}{Agriculture}\/},  {\it
  \bibinfo{volume}{13}\/}, \bibinfo{pages}{628}.
\bibitem[{Wang et~al.(2023{\natexlab{b}})Wang, Li, Nuyttens, Zhang, Liu \&
  Li}]{wang2023evaluationof}
\bibinfo{author}{Wang, S.}, \bibinfo{author}{Li, X.},
  \bibinfo{author}{Nuyttens, D.}, \bibinfo{author}{Zhang, L.},
  \bibinfo{author}{Liu, Y.}, \& \bibinfo{author}{Li, X.}
  (\bibinfo{year}{2023}{\natexlab{b}}).
\newblock \bibinfo{title}{Evaluationof compact air-induction flat fan nozzles
  for herbicide applications: Spray drift and biological efficacy}.
\newblock {\it \bibinfo{journal}{Frontiers in Plant science}\/},  {\it
  \bibinfo{volume}{14}\/}, \bibinfo{pages}{1018626}.

\end{thebibliography}
\nocite{*}







\end{document}